\documentclass[10pt,twocolumn,letterpaper]{article}

\usepackage{wacv}              % To produce the CAMERA-READY version
\usepackage{graphicx}
\usepackage{amsmath}
\usepackage{amssymb}
\usepackage{booktabs}

\usepackage{mathtools}
\usepackage{amsthm}

\usepackage{microtype}
\usepackage{multirow}
\usepackage{float}
\usepackage[dvipsnames]{xcolor}
\usepackage[accsupp]{axessibility} 
\usepackage[pagebackref,breaklinks,colorlinks]{hyperref}

\usepackage[capitalize]{cleveref}
\crefname{section}{Sec.}{Secs.}
\Crefname{section}{Section}{Sections}
\Crefname{table}{Table}{Tables}
\crefname{table}{Tab.}{Tabs.}

\def\wacvPaperID{1440} % *** Enter the WACV Paper ID here
\def\confName{WACV}
\def\confYear{2025}

\begin{document}

% \title{Stylist: Style-Driven Feature Ranking for Robust Novelty Detection}
\title{Robust Novelty Detection through Style-Conscious Feature Ranking}

% \author{Stefan Smeu\thanks{Equal contribution.}\\
% Bitdefender\\
% {\tt\small ssmeu@bitdefender.com}
% \and
% Elena Burceanu\footnotemark[1]\\
% Bitdefender\\
% % First line of institution2 address\\
% {\tt\small eburceanu@bitdefender.com}
% \and
% Emanuela Haller\footnotemark[1]\\
% \\
% % First line of institution2 address\\
% {\tt\small ehaller@bitdefender.com}
% \and
% Andrei Liviu Nicolicioiu\\
% MILA, Montreal \\
% University of Montreal \\ 
% % First line of institution2 address\\
% {\tt\small andrei.nicolicioiu@mila.quebec}
% }

\author{
Stefan Smeu\thanks{Equal contribution.} \textsuperscript{1} ~~~
Elena Burceanu\footnotemark[1] \textsuperscript{1} ~~~
Emanuela Haller\footnotemark[1] \textsuperscript{1} ~~~
Andrei Liviu Nicolicioiu\textsuperscript{2} \\
$^1$Bitdefender, Bucharest, Romania ~~~~~
$^2$Mila and Université de Montréal, Montréal, Canada ~~~~~ \\
% $^3$University of Montreal, Canada \\
{\tt\small\{ssmeu,eburceanu\}@bitdefender.com} ~~
{\tt\small andrei.nicolicioiu@mila.quebec}
}

\maketitle

\begin{abstract}
% Novelty detection aims at finding samples that differ in some form from the distribution of seen samples. But not all changes are created equal. Data can suffer a multitude of distribution shifts, and we might want to detect only some types of relevant changes. Similar to works in out-of-distribution generalization, we propose to use the formalization of separating into semantic or content changes, that are relevant to our task, and style changes, that are irrelevant. Within this formalization, we define the \textbf{robust novelty detection} as the task of finding semantic changes while being robust to style distributional shifts. 
% Leveraging pretrained, large-scale model representations, we introduce {\bf Stylist}, a novel method that focuses on dropping environment-biased features. First, we compute a per-feature score based on the feature distribution distances between environments. Next, we show that our selection manages to remove features responsible for spurious correlations and improve novelty detection performance. For evaluation, we adapt domain generalization datasets to our task and analyze the methods' behaviors. We additionally built a large synthetic dataset where we have control over the spurious correlations degree. We prove that our selection mechanism improves novelty detection algorithms across multiple datasets, containing both stylistic and content shifts.

Novelty detection seeks to identify samples deviating from a known distribution, yet data shifts in a multitude of ways, and only a few consist of relevant changes. Aligned with out-of-distribution generalization literature, we advocate for a formal distinction between task-relevant semantic or content changes and irrelevant style changes. This distinction forms the basis for \textbf{robust novelty detection}, emphasizing the identification of semantic changes resilient to style distributional shifts. To this end, we introduce {\bf Stylist}, a method that utilizes pretrained large-scale model representations to selectively discard environment-biased features. By computing per-feature scores based on feature distribution distances between environments, Stylist effectively eliminates features responsible for spurious correlations, enhancing novelty detection performance. Evaluations on adapted domain generalization datasets and a synthetic dataset demonstrate Stylist's efficacy in improving novelty detection across diverse datasets with stylistic and content shifts. The code is available at \url{https://github.com/bit-ml/Stylist}.
\end{abstract}

\section{Introduction}

In the broader body of literature, Novelty Detection (\textbf{ND}) \cite{kloft2023zero, pimentel2014review, OODsurvey, tack2020csi, ruff2021unifying, yang2021generalized} has conventionally revolved around the identification of notable and meaningful deviations from established data distributions. The ND task is often used interchangeably with the broader anomaly detection task, but there is a notable difference between the two. Anomalies are fundamentally distinct from typical samples and can manifest as deviations in various forms. Novelties, or semantic anomalies, represent a subset of anomalies, specifically targeting semantic deviations, aiming to identify any test sample that does not conform to the established training categories. For example, in practical scenarios such as medical diagnosis \cite{Chauhan2015AnomalyDI}, financial fraud detection \cite{journals/dss/BhattacharyyaJTW11} or network intrusion detection \cite{anoshift}, the primary objective is to detect novelties, such as unique aspects of a cell's biological structure, while disregarding irrelevant divergent characteristics, such as artifacts stemming from equipment. 
% poate sa aducem exemplul cu self-driving car de mai jos aici?

Our main point is that not all changes are created equal. When we move across a continent using a self-driving car, we might be amazed by the style of different houses that we have not seen before, but the self-driving car should still behave the same. On the other hand, when encountering a new structure that it has not seen before, such as a new type of intersection or bridge, the car should \textit{detect} that this is a \textit{novel} situation and cease the control to the driver.  

Thus, we define \textbf{semantic} or \textbf{content} shifts as the changes in data distribution that involve factors relevant to our task (such as driving), and \textbf{style} shifts as the changes that involve some factors that are irrelevant to our task
%(see Fig.~\ref{fig:setup_figure})
. In many cases, the style factors are correlated with content factors, so when learning the semantics of a problem, we might learn some spurious correlations involving irrelevant style factors. These spurious correlations might not always hold; thus, we should not rely on them. In this context, we focus on \textbf{robust novelty detection}, which aims to identify distribution changes in content while ignoring style changes. 

To distinguish between the two, we consider the multi-environment setup from the distribution shift studies \cite{koh2021wilds, zhou2022domain}, where, besides the usual content label, we also have access to a style label. An environment is composed of samples with a particular style category, but with any content categories. In this scenario, a style category is essentially a set of factors or relations that hold only in one environment (\eg for the self-driving car example, driving in the forest, near a beach, or even in some fictional, Disney-like scenario can be seen as different styles). On the other hand, a content category refers to a set of factors or relations that hold across all environments (\eg roads, cars, bikes, human categories). The style component characterizes the data in an uncertain, maybe even spurious way, toward the content classification task. During training, the content may be correlated with other factors from the training environments, which are irrelevant to this new task and might become spurious. This is a challenging problem for content classification tasks and even more challenging in the novelty detection setup, where, during training, you only observe a set of known classes.

With this in mind, our work centers on detecting novel content, while removing environment-biased features. Specifically, we propose a method to rank features based on their distributional changes across training environments. This ranking mechanism, followed by the removal of environment-biased features, aims to enhance the performance of novelty detection methods, enabling them to generalize more effectively in the presence of spurious correlations and providing insights into the interpretability of features.

\begin{figure}[t!]
    % \begin{subfigure}{0.5\textwidth}
    %     \centering
    % \includegraphics[width=\columnwidth]{figures2/setup_1.pdf}
    % \end{subfigure}
    \begin{subfigure}{0.5\textwidth}
        \centering
    \includegraphics[width=\columnwidth]{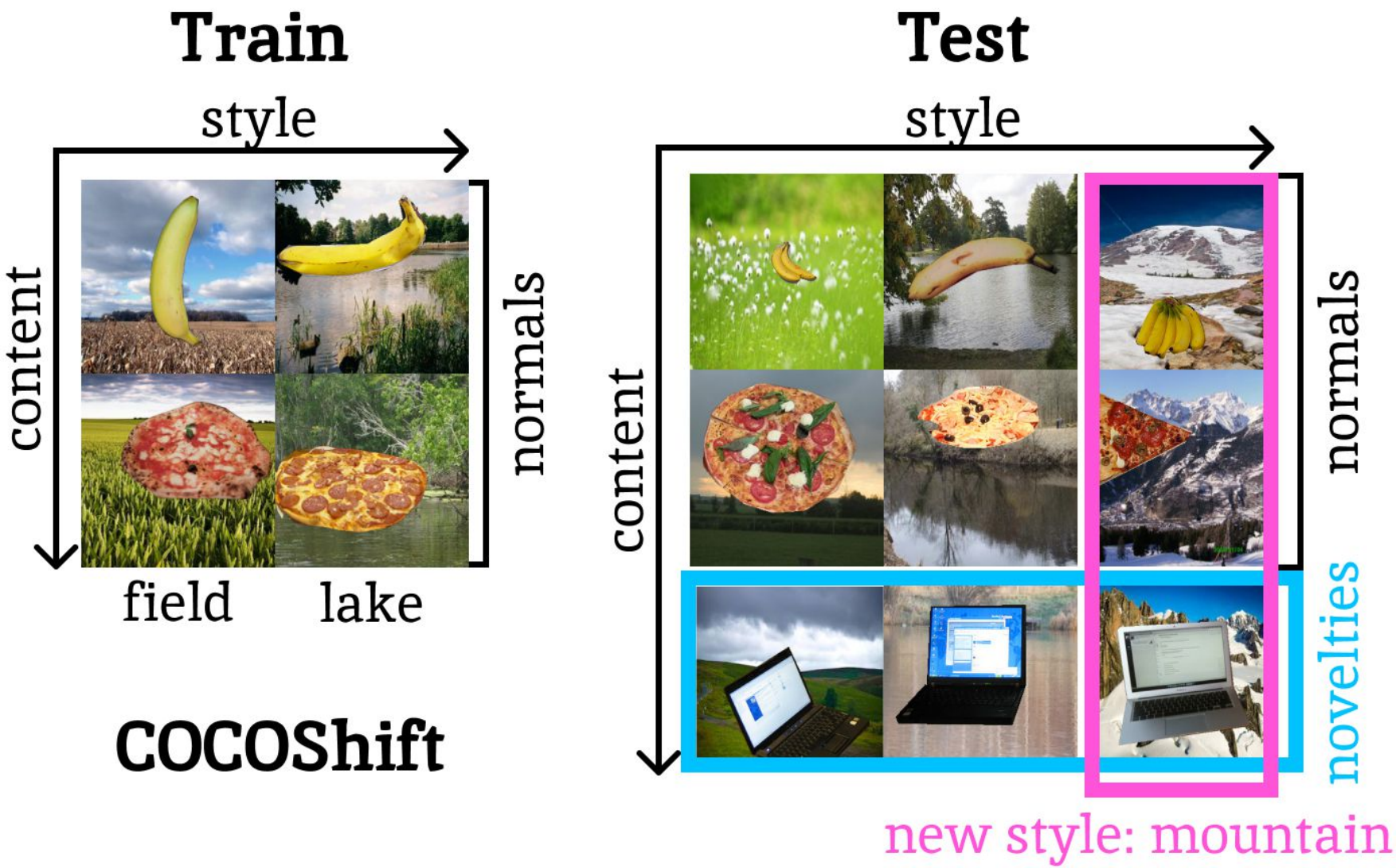}
    \end{subfigure}
    \caption{Multi-env setup for the Robust Novelty Detection task.}
    \label{fig:setup_figure}
\end{figure}

Our \textbf{main contributions} are the following:\\
% \noindent\textbf{ 1. } We transformed existing methods that are traditionally used for feature selection to be environment-aware. On top, we introduce a simple, yet very effective algorithm, \textbf{Stylist}, that scores pretrained features, based on their distribution changes between training environments. We empirically and explicitly prove that this approach \textbf{ranks features based on how much they are focused on the environmental details} and gives a glimpse of interpretability to the "black-box" embeddings. 
\noindent\textbf{ 1. } We show that feature selection based on environment information helps to detect novelties in the presence of irrelevant changes, a setup we call \textbf{Robust Novelty Detection}.\\
% We transformed existing methods that are traditionally used for feature selection to be environment-aware. On top, 
\noindent\textbf{ 2. }
We introduce a simple, yet highly effective algorithm, \textbf{Stylist}, that scores pretrained features, based on their distributional changes between training environments. We empirically % and explicitly 
prove that it \textbf{ranks features based on how much they focus on environmental details} and gives a glimpse of interpretability to the "black-box" embeddings. \\
\noindent\textbf{ 3. } We show that, by gradually removing the environment-biased features proposed by Stylist, we significantly improve the ND models' generalization capabilities, both in the covariate and sub-population shift setups, by up to 8\%.%. with the benefits becoming more pronounced at a high degree of spuriousness correlations.

\noindent\textbf{ 4. } We introduce \textbf{COCOShift}, a comprehensive, synthetic benchmark%, with 4 levels of spuriousness, 
which enables a detailed analysis for the Robust Novelty Detection. We also adapt the DomainNet and fMoW multi-environment real datasets to novelty detection and validate our main results in this setting.

% \begin{figure*}[t!]
%     \begin{subfigure}[t]{0.5\textwidth}
%         \centering
%         \includegraphics[width=\textwidth]{figures/main_figure_a.pdf}
%     \end{subfigure}
%     \begin{subfigure}[t]{0.5\textwidth}
%         \centering
%         \includegraphics[width=\textwidth]{figures/main_figure_b.pdf}
%     \end{subfigure}
%     \caption{{\bf Stylist}. We improve the ND performance by identifying (Step 1) and gradually removing (Step 2) environment-biased features. From this point of view, higher distribution distances between environments proved to be a good indicator for ranking features.}
%     \label{fig:main_figure}
% \end{figure*}

\begin{figure*}[t!]
    \begin{subfigure}{0.5\textwidth}
        \centering
    \includegraphics[width=0.9\columnwidth]{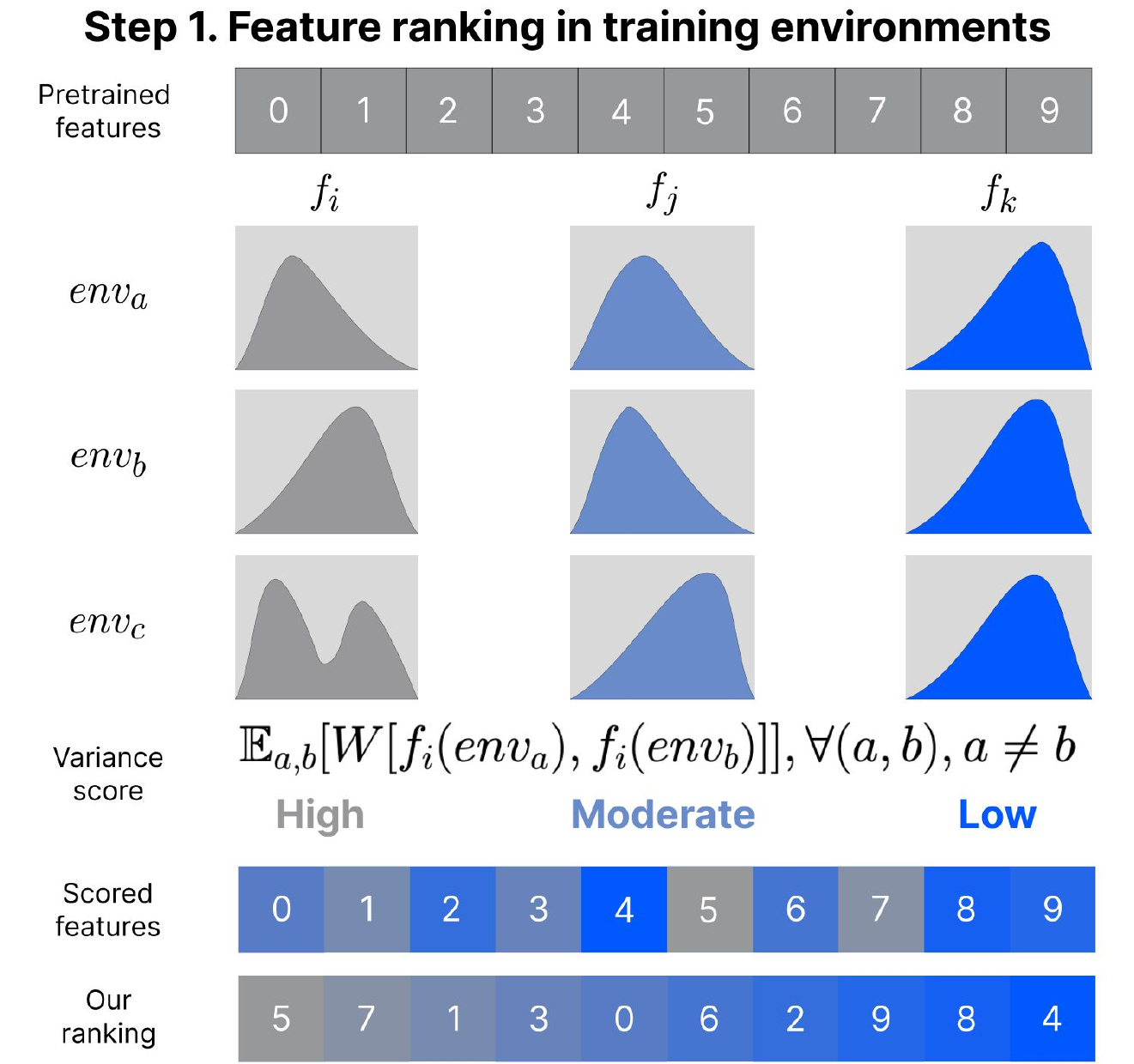}
    \end{subfigure}
    \begin{subfigure}{0.5\textwidth}
        \centering
    \includegraphics[width=0.9\columnwidth]{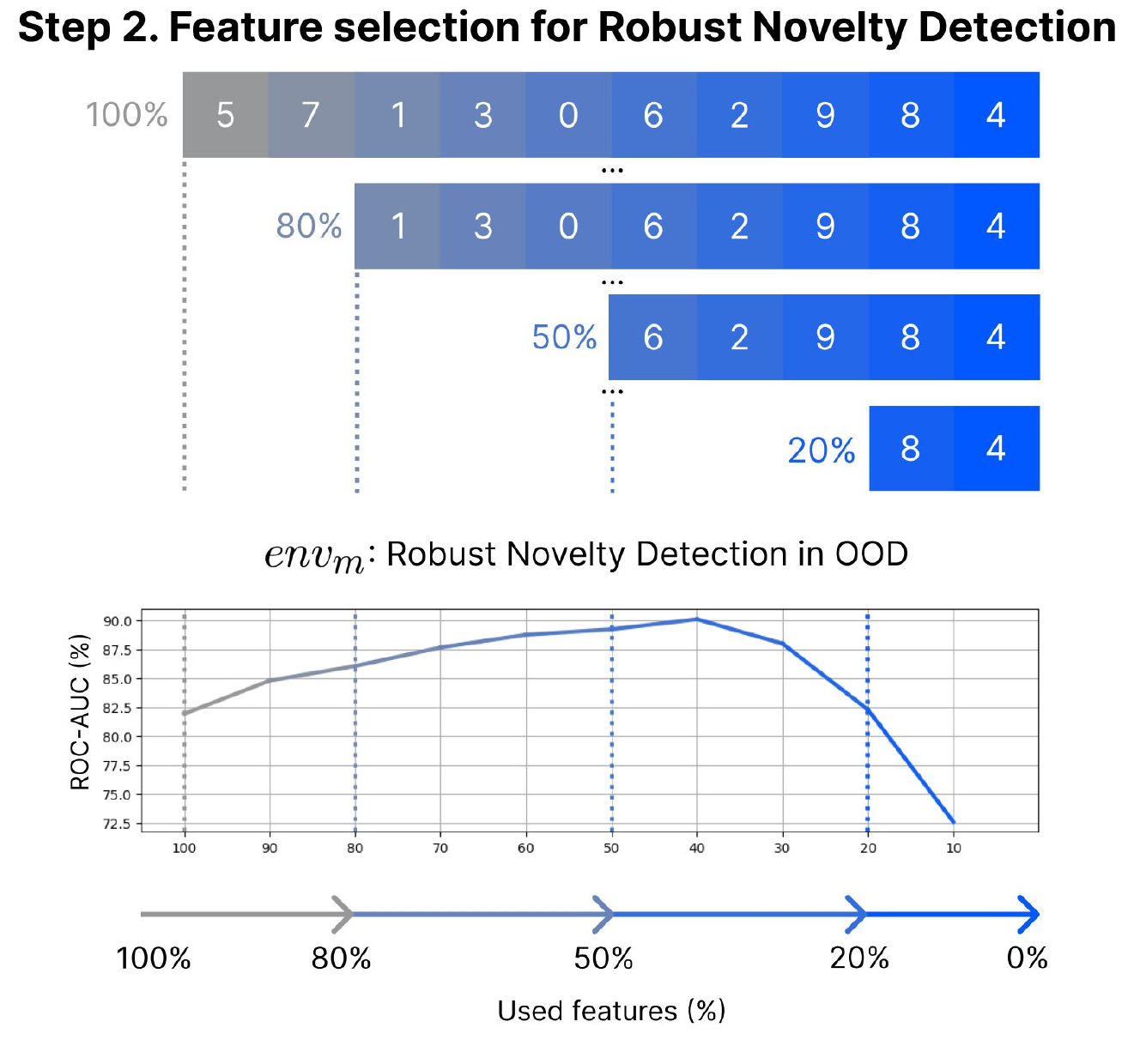}
    \end{subfigure}
    \caption{{\bf Stylist}. We improve the ND performance by identifying (Step 1) and gradually removing (Step 2) environment-biased features. From this point of view, higher distribution distances between environments proved to be a good indicator for ranking features.}
    \label{fig:main_figure}
\end{figure*}

\section{Problem formulation}

Real-world data suffers from a multitude of changes that we usually refer to as distributional shifts. As described by~\cite{ea_moco} these kinds of shifts are involved in different lines of work, with different goals: domain generalization wants to be \textit{robust to style shifts} while most anomaly detection methods want to \textit{detect either style or semantic shifts}. We denote robust novelty detection as the task of detecting semantic novelties while being robust to style distributional shifts. More exactly, detect samples that differ by some semantic shifts from some \textit{seen} training samples, while ignoring samples that are only affected by style shifts.

We work in a multi-environment setup, where each training environment changes the style of the samples, while all environments contain a set of \textit{seen} content classes. The goal of training environments is to define what is content and what is style. Consequently, we are not restricted to a certain definition of style, but rather expect the training environments to define what might be correlated with our task, but is not actually relevant. Then, we define an evaluation environment, containing \textit{both seen and \textbf{novel}} classes with an associated new style. The goal of \textbf{robust novelty detection} is to separate between \textit{seen} and \textit{novel} content classes, without being affected by the new style. 

We focus on multi-class novelty detection, where we have training environments with multiple content classes. However, we treat them as a single group of normal samples and ignore their content labels. By the definition of novelty detection task, there is a zero level of corruption among the normal samples, as opposed to the more common setup of anomaly detection.   

In Fig.~\ref{fig:setup_figure} we present two scenarios to exemplify our setup. In the first example, normal samples encompass representations of objects in various formats (real images or paintings). In this context, style is defined as the manner of depiction. During testing, our objective is to correctly categorize the laptop as a novel class. Furthermore, we must discern that the sketch of the banana, despite the shift in style (from real images and paintings to sketches), is not a novel class. In the second example, we observe a different definition of style, namely the background of the images, which should also be irrelevant for classifying the content.

\section{Our approach}

%We operate on top of pretrained embeddings with extensive knowledge of a multitude of content and style categories. 
Some dimensions of a given pretrained representation could be more representative of the semantic aspects, while others might be more representative of style elements. To minimize the impact of style factors on our novelty detection task, we aim to reduce our reliance on them. Thus, it might be that we are better off ignoring the dimensions that mostly contain style information, which we denote as environment-biased features. We focus on discovering which features from a given, pretrained representation, are more environment-biased, thus prone to contain spurious correlations, and should be better ignored. Finding the robust part of a representation is closely linked to invariance between environments, thus we want to have a measure of variance for each dimension in our representation. We first quantify the degree of change in each feature distribution, and then we drop the ones that vary more, as depicted in Fig.~\ref{fig:main_figure}.

We assume that for each sample of our training set, we start with a vector of $N$ features, extracted from a pretrained model. We proceed in two steps:

\paragraph{Step 1. Feature ranking in training environments} First, we compute a score that says how much a feature changes across environments. For each feature $i$, we consider $f_i(env)$ to be the distribution of this feature in environment $env$. In our case, we use the feature histogram per environment.

\begin{equation}
f_i(env) = p(f_i|env), \quad \forall i \in [1..N], \forall env \in all_{.}envs.
\label{eq:f_distr}
\end{equation}

Next, we employ the Wasserstein distance to compute the distance between the distributions of each feature $i$, across pairs of environments $(a, b)$.

\begin{equation}
dist_i(env_a, env_b) = W(f_i(env_a), f_i(env_b)), \quad \forall i \in [1..N] 
\label{eq:feature_ranking_1}
\end{equation}

The per-feature score is then obtained as the expected value of the Wasserstein distance across all pairs of environments $(env_a, env_b)$, where $a \neq b$.

\begin{equation}
 \begin{split}
score_i = \mathbb{E}_{a,b}[dist_i(env_a, env_b)] , \quad \forall i \in [1..N],\\\forall \text{ training } env_a \neq env_b,
\end{split}
\label{eq:feature_ranking_2}
\end{equation}

\paragraph{Step 2. Features selection for Robust Novelty Detection} Next, since our purpose is to be robust and to be able to ignore environmental changes, we remove features with the highest scores. The intuition here is that environment-biased features facilitate spuriousness, providing a training setup prone to such correlations.

The exact distance used might not be that important, but what matters is the process of looking at differences between environments and searching for what consistently changes between them (\eg in terms of distribution). For this, in our approach, we rely on the following assumptions, which we argue that are not very restrictive, but rather grounded in common sense:
% \begin{enumerate}
%     \item The pretrained feature extractor is able to represent both seen and novel content categories, as well as known and new styles. 
%     \item The change in style dominates over the change in content, when we compare between different environments.
% \end{enumerate}
\paragraph{1. The feature extractor} It is mandatory for our feature extractor to "see" both known and new content, but also styles. Missing discriminative features between new and known content, makes our task impossible to approach. On the other hand, having features that are non-discriminative of style, makes the robust ND task useless, since there is no information related to the style that the algorithm needs to adapt to ignore. This assumption is easily met in practice nowadays, when we have access to powerful pretrained models that have been trained on large and comprehensive datasets. Thus, the difficulty does not lie in getting good representation, but at the next level, where, given a set of very descriptive features, you need to select the ones that are relevant for identifying novel content, while dropping style-related features that can cause spurious correlations. A clarifying example for motivating the need for this assumption and its relevance is the following: "Alert me when you see wild animals, intruding into my garden, engaging with my pets, or farm animals". In this case, the ND task could be to detect if something abnormal - wild - appears, after training the model with a collection of normal images.

% First, our approach leverages pretrained embeddings with extensive coverage across various content and style categories. This eliminates the need for employing domain adaptation techniques on the pretrained representation, enabling us to keep the model frozen and exclusively utilize its features. To illustrate this concept, consider the initial scenario presented in Fig.~\ref{fig:setup_figure}, where we presume that the pretrained model effectively captures pertinent features for objects like bananas, pizza, and laptops across diverse depictions such as real images, paintings, and sketches. This assumption is easily met in practice nowadays when we have access to powerful pretrained models that have been trained on large and comprehensive datasets. So the difficulty does not lie in getting good representation, but at the next level, where, given a set of features, you need to select the ones that are relevant for identifying novel content while dropping style-related features that can cause spurious correlations. 

\paragraph{2. Style changes more between environments} In our setup, both style and content can vary across environments. We assume that style-induced changes in the data distribution are greater than content-induced ones, when we look at two different environments. Hence, if the style is changing more, the content is changing less, and a natural interpretation of this assumption is that class distribution across environments is similar. While our method clearly benefits from such a setup, this is not a hard assumption we need. In our experiments for ND, we have two classes (normal vs novel), that aggregates over multiple real content one. Those real classes are usually very heterogeneous, covering even $340$ for some datasets (see Appendix~\ref{appendix:benchmarks}), completely disregarding the interpretation that the content should be similarly distributed across environments for Stylist to work.

% As for the second assumption, in our setup, both style and content can vary across environments. Since random changes would create an impossible problem, we assume that style-induced changes in the data distribution are greater than content ones, when we look at two different environments. This implies that our environments have different associated styles, and the considered representation can capture these relevant differences, which is covered by our first assumption. In order to mitigate these challenges, it is imperative that we operate in a well-constructed multi-environment setup to be able to gain relevant information about the style shift.  In the context of the Novelty Detection task, where both style and content distributions can shift, the quality and quantity of these environments play a pivotal role. They serve as reference points to delineate and clarify what holds significance and what does not within our task.

\section{Experimental analysis}

\begin{table*}[t!]
    \centering
    \caption{\textbf{Novelty Detection Methods on top of Stylist features}. Notice how, for almost all ND algorithms and dataset combinations, dropping top environment-biased features, as identified by Stylist, increases the ROC-AUC performance (see the improvement in green).}
    \begin{tabular}{l ccc ccc ccc}
        \toprule
        \multirow{5}{*}{\shortstack{\bf Novelty Detection \\ \bf Method}} & \multicolumn{3}{c}{\textbf{fMoW}} & \multicolumn{3}{c}{\textbf{DomainNet}} & \multicolumn{3}{c}{\textbf{COCOShift95}} \\
        \cmidrule(lr){2-4}
        \cmidrule(lr){5-7}
        \cmidrule(lr){8-10}
        & \multicolumn{2}{c}{ROC-AUC $\uparrow$} & \multirow{2}{*}{\shortstack{\% \\ selected \\ feat.}} & \multicolumn{2}{c}{ROC-AUC $\uparrow$} & \multirow{2}{*}{\shortstack{\% \\ selected \\ feat.}} & \multicolumn{2}{c}{ROC-AUC $\uparrow$} & \multirow{2}{*}{\shortstack{\% \\ selected \\ feat.}} \\
        \cmidrule(lr){2-3}
        \cmidrule(lr){5-6}
        \cmidrule(lr){8-9}
        & \shortstack{all \\ feat.} & \shortstack{\textbf{Stylist} \\ feat.} & & \shortstack{all \\ feat.} & \shortstack{\textbf{Stylist} \\ feat.} & & \shortstack{all \\ feat.} & \shortstack{\textbf{Stylist} \\ feat.} \\
        \toprule
        OCSVM & 46.9 & 54.3 {\footnotesize{\textcolor{ForestGreen}{(+7.4)}}} & 85 & 50.4 & 51.4 {\footnotesize{\textcolor{ForestGreen}{(+1.0)}}} & 95 & 52.6 & 58.4 {\footnotesize{\textcolor{ForestGreen}{(+5.8)}}} & 90 \\
        LOF & 58.0 & 60.8 {\footnotesize{\textcolor{ForestGreen}{(+2.8)}}} & 15 & 51.1 & 52.0 {\footnotesize{\textcolor{ForestGreen}{(+0.9)}}} & 90 & 83.4 & 86.5 {\footnotesize{\textcolor{ForestGreen}{(+3.1)}}} & 30 \\
        kNN & 59.0 & 60.3 {\footnotesize{\textcolor{ForestGreen}{(+1.3)}}} & 20 & 50.6 & 50.8 {\footnotesize{\textcolor{ForestGreen}{(+0.2)}}} & 40 & 79.8 & 85.1 {\footnotesize{\textcolor{ForestGreen}{(+5.3)}}} & 30 \\
        kNN norm & 41.9 & 49.9 {\footnotesize{\textcolor{ForestGreen}{(+8.0)}}} & 5 & 52.5 & 52.8 {\footnotesize{\textcolor{ForestGreen}{(+0.3)}}} & 70 & 86.2 & 86.2 {\footnotesize{\textcolor{ForestGreen}{(+0.0)}}} & 100\\
        \cmidrule(lr){1-10}
        kNN+ & 58.0 & 60.8 {\footnotesize{\textcolor{ForestGreen}{(+2.8)}}} & 15 & 51.1 & 52.0 {\footnotesize{\textcolor{ForestGreen}{(+0.9)}}} & 90 & 82.3 & 82.3 {\footnotesize{\textcolor{ForestGreen}{(+0.0)}}} & 100\\
        \bottomrule
    \end{tabular}
    \label{tab:sota_novelty_detection}
\end{table*}

Our experimental analysis is conducted using two real datasets and a synthetic one. For the first two, we employ adaptations of well-established domain generalization datasets: fMoW~\cite{fmow} and DomainNet~\cite{DomainNet}. All are multi-environment datasets and for each, we divide the environments into two sets denoted as follows: in-distribution (\textbf{ID}) environments (associated with styles that we observe during training) and out-of-distribution (\textbf{OOD}) environments (associated with styles that we only observe during testing). Each dataset contains a set of annotated semantic categories, and we divide them into two sets: {\bf normal} classes (content categories observed during training) and {\bf novel} classes (content categories that should be distinguished from normal ones during testing). For each sample, we have a style label and a novelty label (normal vs. novel).  

%The general idea is to use the environments present in the dataset to separate the stylistic differences into "in-distribution"(ID) and "out-of-distribution"(OOD) splits. We borrow the terminology from the domain generalization literature, where the "in-distribution" split is used for training, while the "out-of-distribution" is used for testing, denoting a shift in style. Simultaneously, we assign some of the classification targets (classes) for training and declare the others as novelties to be detected. 

\textbf{fMoW} comprises satellite images of various functional buildings. The style is defined by the location of the image, while the content is defined by the class of the observed building. To generate a greater shift between ID and OOD styles, we considered photos taken in Europe, America, Asia, and Africa to compose the ID environments, while those taken in Australia were used as OOD ones. The content separation into normal and novel categories was randomly generated (see Appendix \ref{appendix:fmow}).

\textbf{DomainNet} contains images of common objects in six different domains. The style is defined by the domain, while the content is defined by the object class. We separated the environments into ID: clipart, infograph, painting, and real and OOD: quickdraw and sketch. We randomly split the classes into normals and novelties (see Appendix \ref{appendix:domainnet}).

\textbf{COCOShift} is a synthetic dataset generated to allow an in-depth analysis of our approach. We combined segmented objects from COCO~\cite{COCO} with natural landscape imagery from Places365~\cite{Places365}. The landscape images define the style of the data, while object categories depict the content. We have grouped the landscape images into 9 categories (\eg forest, mountain), each of an equal number of samples, and further split them into 5 ID and 4 OOD styles. The object categories were split into normal and novel classes by following a proper balancing between them. (see Appendix \ref{appendix:cocoshift}). \textbf{Spuriousness}: For COCOShift, we deliberately introduced and varied the level of spurious correlations between style and content, similar to~\cite{DFR, DRO}. The spuriousness level ranges from 50\% (balanced dataset, without spurious correlations) to 95\% (where the normal class is strongly correlated with some environments, while we observe few samples in the rest of the environments). We obtain the COCOShift benchmark, with 4 levels of spuriousness in the training sets: COCOShift\_balanced, COCOShift75/90/95. %COCOShift75, COCOShift90, and COCOShift95. 

\textbf{Metrics}: For our ND experiments, we report the ROC-AUC metric, as the average performance over test environments. Unless otherwise specified, we report performance over OOD environments. 

\textbf{Feature selection algorithms}: We have transformed InfoGain~\cite{infogain} and FisherScore~\cite{fisher} to identify and then discard the environment-biased features. Along with our Stylist method, we denote those three methods as \textit{'Env-Aware'} methods. As \textit{'Not Env-Aware'} methods, we evaluate MAD (mean absolute difference), Dispersion (as the ratio between arithmetic and geometric mean), Variance, and PCA Loadings. We use all those methods to compute an individual score per feature (see Appendix~\ref{apx:selection_algo_ablation} for details). 

\noindent\textit{Env-InfoGain}: We compute the mutual information between each feature and the style labels. High scores indicate a higher dependency between feature and style labels.

\noindent\textit{Env-FisherScore}: We rank the features based on their relevance for the classification of style categories. 

\textbf{Novelty detection algorithms}: We observe the impact of our method on several ND solutions: OCSVM~\cite{ocsvm}, LOF~\cite{lof}, and kNN~\cite{knn} with different variations (normalized or not at sample level, with 10 or 30 neighbors to measure variations). We also tested the impact in the state-of-the-art solution for OOD detection, kNN+~\cite{fast-knn-ood}, which trains a kNN on top of normalized samples, but on top of ResNet-18 features, fine-tuned using a supervised contrastive loss like in \cite{supcon}.

\textbf{Pretrained features}: We validate over multiple feature extractors, from different tasks, architectures, and datasets (supervised, multi-modal, contrastive, from basic ResNet to Visual Transformers, trained on ImageNet~\cite{imagenet} and other larger datasets): ResNet-18, ResNet-34~\cite{resnet}, CLIP~\cite{clip}, ALIGN~\cite{align}, BLIP-2~\cite{blip2}. Unless otherwise specified, the experiments use ResNet-18, pretrained on ImageNet.

\subsection{Robust Novelty Detection}

\begin{figure*}[t]
    \begin{subfigure}{0.33\textwidth}
        \centering
        \includegraphics[width=1\linewidth]{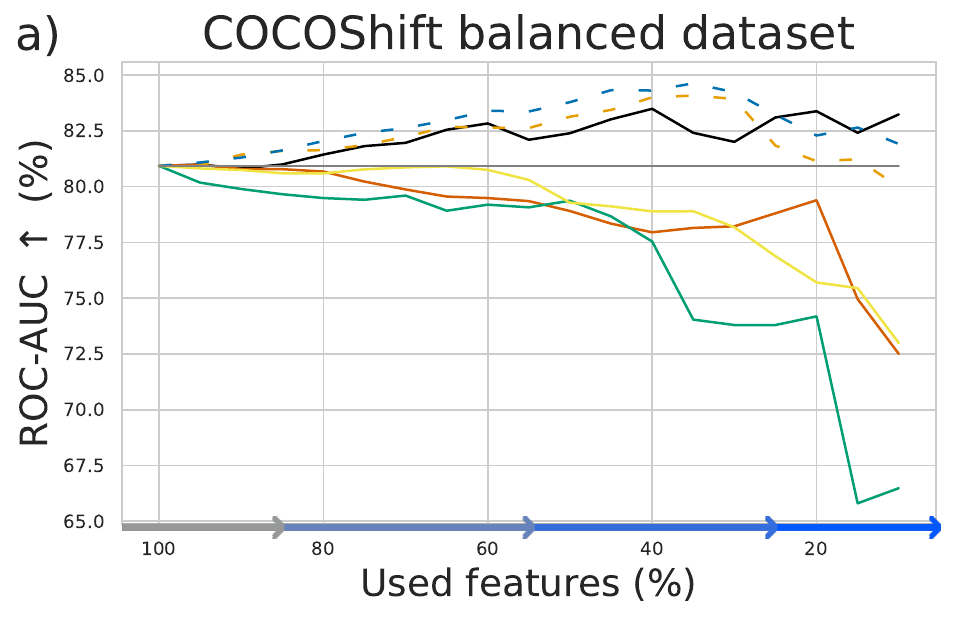}
    \end{subfigure}
    \begin{subfigure}{0.33\textwidth}
        \centering
        \includegraphics[width=1\linewidth]{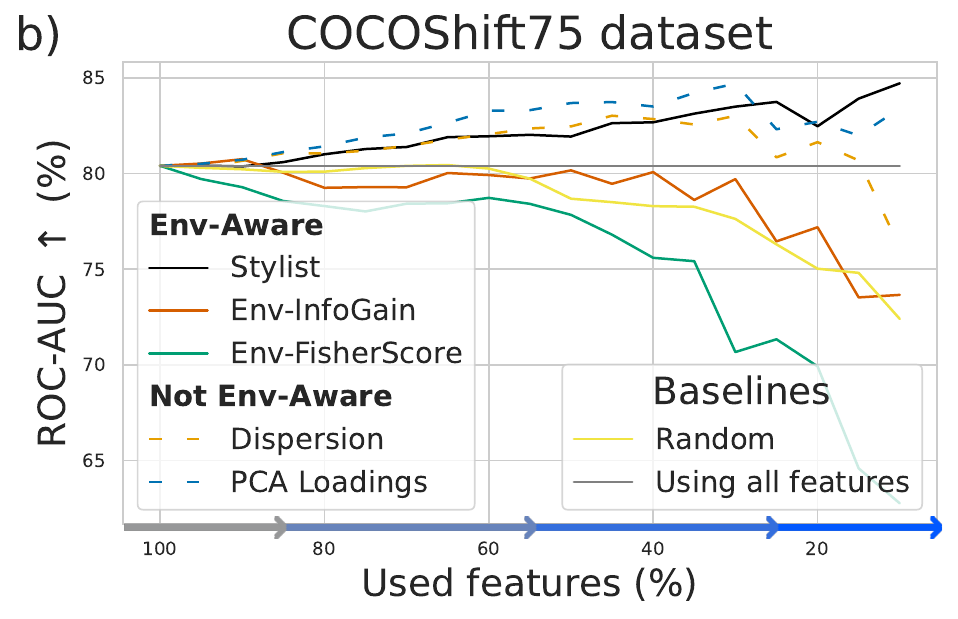}
    \end{subfigure}
    \begin{subfigure}{0.33\textwidth}
        \centering
        \includegraphics[width=1\linewidth]{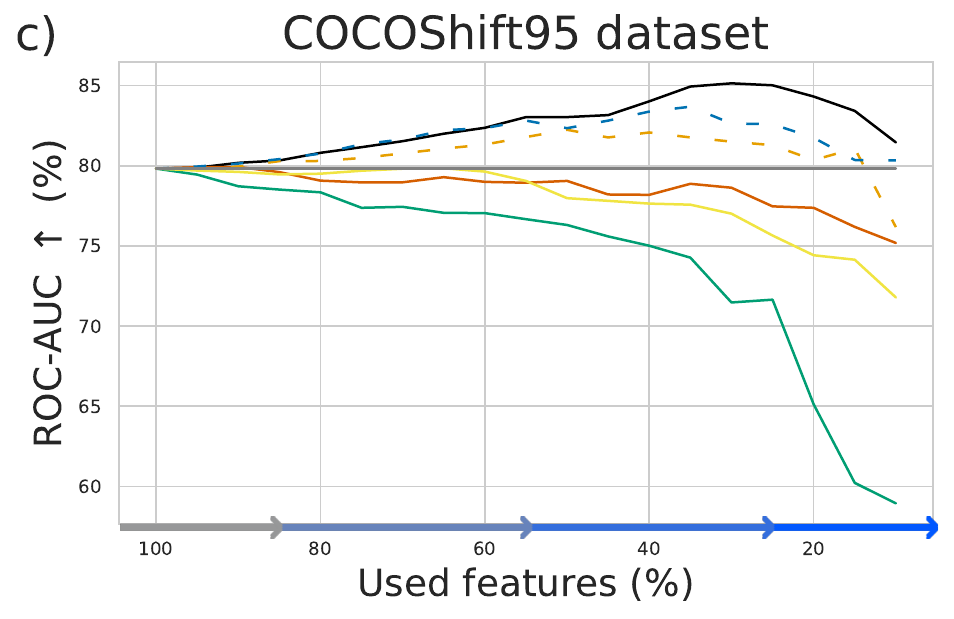}
    \end{subfigure}
    \caption{\textbf{Feature selection algorithms}. From left to right on the horizontal axis, we remove features according to the ranking of each feature selection algorithm. As the spuriousness level of the train set increases ($a) \rightarrow b) \rightarrow c)$), the performance of Stylist (in black) increases, while the performance of other methods decreases. This proves that our approach is better at identifying environment-biased features responsible for the spurious correlations. The reported ROC-AUC performance is for the same OOD sets in all three plots.} 
    \label{fig:sota_feature_selection}
\end{figure*}

\begin{figure*}[t!]
    \begin{subfigure}{0.33\textwidth}
        \centering
        \includegraphics[width=\linewidth]{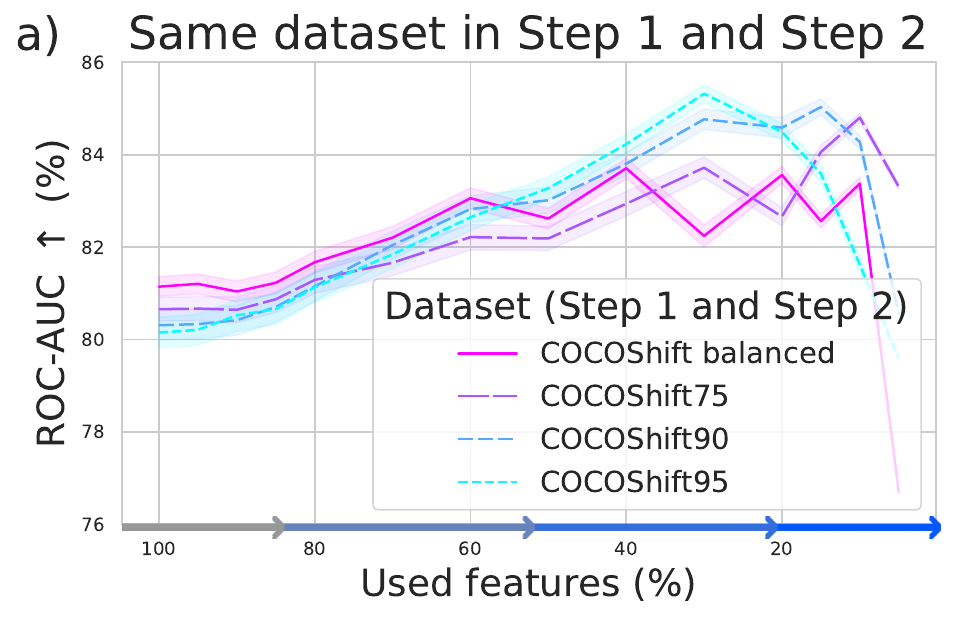}
    \end{subfigure}
    \begin{subfigure}{0.33\textwidth}
        \centering
        \includegraphics[width=\linewidth]{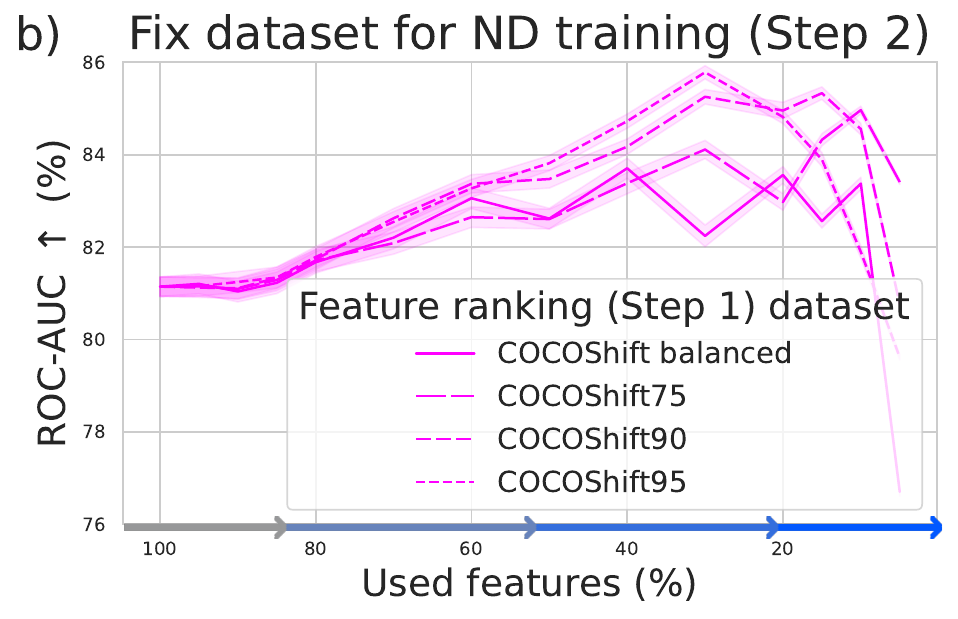}
    \end{subfigure}
    \begin{subfigure}{0.33\textwidth}
        \centering
        \includegraphics[width=\linewidth]{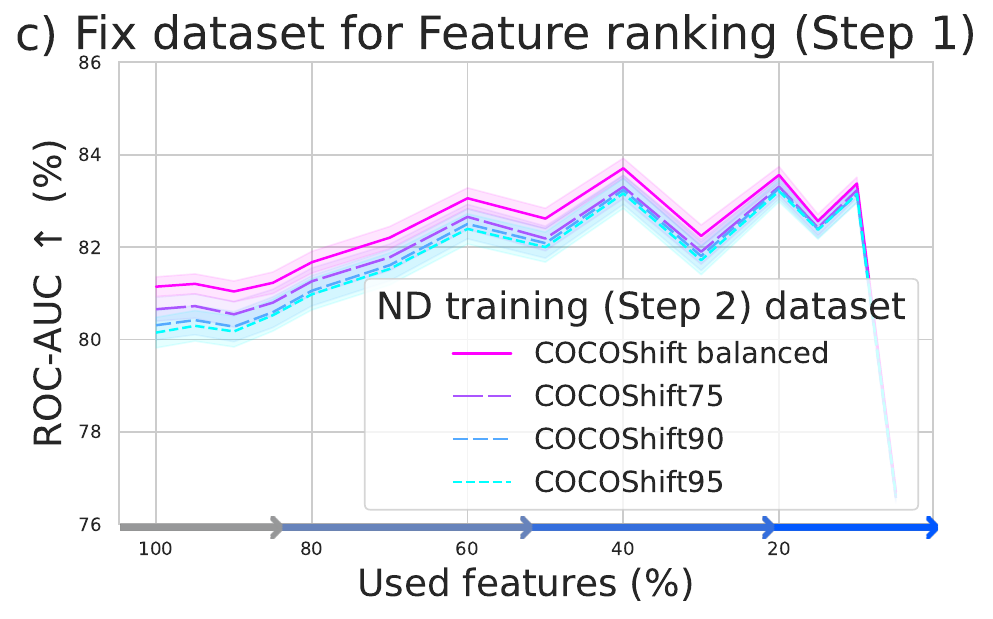}
    \end{subfigure}
    \caption{\textbf{Dataset spuriousness impact}. We vary the train set spuriousness level between style and content for the two steps of our algorithm. {\bf a)} same dataset for both steps; {\bf b)} fixed dataset (COCOShift\_balanced) for ND training in Step 2; {\bf{c)}} fixed dataset (COCOShift\_balanced) for Feature ranking in Step 1.  Our method always manages to improve the ND performance (w.r.t. all features baseline), even in degenerated cases like 95\% (or no) spurious correlation, in only one or in both steps (see the positive slopes in all curves).}
    \label{fig:spuriousness}
\end{figure*}
\noindent\textbf{Stylist for Novelty Detection} We evaluate how our selection affects the robustness of various Novelty Detection algorithms. Tab.~\ref{tab:sota_novelty_detection} presents the initial performance using all features and the best result achieved by dropping environment-biased features as identified by {\it Stylist}. Notice how for almost all cases, using only a percentage of features improves performance by up to 8\%.

\noindent\textbf{Comparison with other feature selection algorithms} % de ce; ce; cum; rezultate
We compare in Fig.~\ref{fig:sota_feature_selection} between different methods of feature selection. For all algorithms, we drop features ranked as the most irrelevant. We see that, as we vary the spuriousness level in the training dataset, the relative order of the algorithms changes, showing that some perform better when working on a balanced dataset (like PCA based ones), while our {\it Stylist} works the best in difficult scenarios with an increased level of spurious correlations. 
Please refer to Appendix~\ref{apx:selection_algo_ablation} for the individual performances and notice in Appendix~\ref{apx:sota_sub_population} how those covariate shift results are consistent even for the sub-population OOD shifts. Also, Appendix \ref{appendix:qualitative} shows a qualitative analysis. 
% This main experiment in our work focuses on the impact of our feature ranking for \textbf{robust novelty detection}. For this, we gradually remove what we expect to be features that are more environment-biased, according to our ranking. Next, we use the rest of the features as input in ND algorithms. We see in Fig.~\ref{fig:nd_coco_dn_no_corr} that while dropping features, the performance improves for a while and then drops when we are left with significantly fewer features (that might be related to the label). This validates our assumption that our algorithm gives higher scores to environment-specific features.
% Moreover, we show the results are consistent both for covariate and sub-population OOD shifts. 

\begin{table*}[t!]
    \centering
    \caption{\textbf{Feature extractors}. Stylist improves the performance for all types of pretrained features considered, over all three datasets. For simplicity, we use only ResNet-18 in other experiments.}
    \setlength{\tabcolsep}{6pt}
    \begin{tabular}{l ccc ccc ccc}
        \toprule
        \parbox[t]{0mm}{\multirow{5}{*}{\bf Features}}& \multicolumn{3}{c}{\textbf{fMoW}} & \multicolumn{3}{c}{\textbf{DomainNet}} & \multicolumn{3}{c}{\textbf{COCOShift95}} \\
        \cmidrule(lr){2-4}
        \cmidrule(lr){5-7}
        \cmidrule(lr){8-10}
        & \multicolumn{2}{c}{ROC-AUC $\uparrow$} & \multirow{2}{*}{\shortstack{\% \\selected\\feat.}} & \multicolumn{2}{c}{ROC-AUC $\uparrow$ } & \multirow{2}{*}{\shortstack{\%\\selected\\feat.}} & \multicolumn{2}{c}{ROC-AUC $\uparrow$ } & \multirow{2}{*}{\shortstack{\%\\selected\\feat.}}\\
        \cmidrule(lr){2-3}
        \cmidrule(lr){5-6}
        \cmidrule(lr){8-9}
        & \shortstack{all\\feat.} & \shortstack{\textbf{Stylist}\\ feat.} & &  \shortstack{all\\feat.} & \shortstack{\textbf{Stylist}\\ feat.} & & \shortstack{all\\feat.} & \shortstack{\textbf{Stylist}\\ feat.}\\ 
        \toprule
        ResNet-18 & 59.0 & 60.3 \footnotesize{\textcolor{ForestGreen}{(+1.3)}} & 20 &50.6 & 50.8 \footnotesize{\textcolor{ForestGreen}{(+0.2)}} & 40 &79.8 & 85.1 \footnotesize{\textcolor{ForestGreen}{(+5.3)}} & 30 \\
        ResNet-34 & 61.9 & 65.6 \footnotesize{\textcolor{ForestGreen}{(+3.7)}} & 30 & 51.1 & 51.1 \footnotesize{\textcolor{ForestGreen}{(+0.1)}} & 40 & 78.9 & 82.6 \footnotesize{\textcolor{ForestGreen}{(+3.7)}} & 20\\
        CLIP &  54.3 & 55.5 \footnotesize{\textcolor{ForestGreen}{(+1.3)}} & 25 & 60.8 & 61.5 \footnotesize{\textcolor{ForestGreen}{(+0.8)}} & 30 & 94.5 & 94.9 \footnotesize{\textcolor{ForestGreen}{(+0.4)}} & 95 \\
        ALIGN & 54.6 & 56.2 \footnotesize{\textcolor{ForestGreen}{(+1.6)}} & 40 &60.6 & 60.8 \footnotesize{\textcolor{ForestGreen}{(+0.3)}} & 75 & 89.6 & 89.7 \footnotesize{\textcolor{ForestGreen}{(+0.1)}} & 80 \\
        BLIP-2 & 58.6 & 59.1 \footnotesize{\textcolor{ForestGreen}{(+0.4)}} & 15 & 65.1 & 65.8 \footnotesize{\textcolor{ForestGreen}{(+0.7)}} & 20 & 96.7 & 96.8 \footnotesize{\textcolor{ForestGreen}{(+0.1)}} & 95\\
        \bottomrule
    \end{tabular}
    \label{tab:feature_extractors}
\end{table*}

\begin{figure*}[t!]
    \begin{subfigure}{0.33\textwidth}
        \centering
        \includegraphics[width=\linewidth]{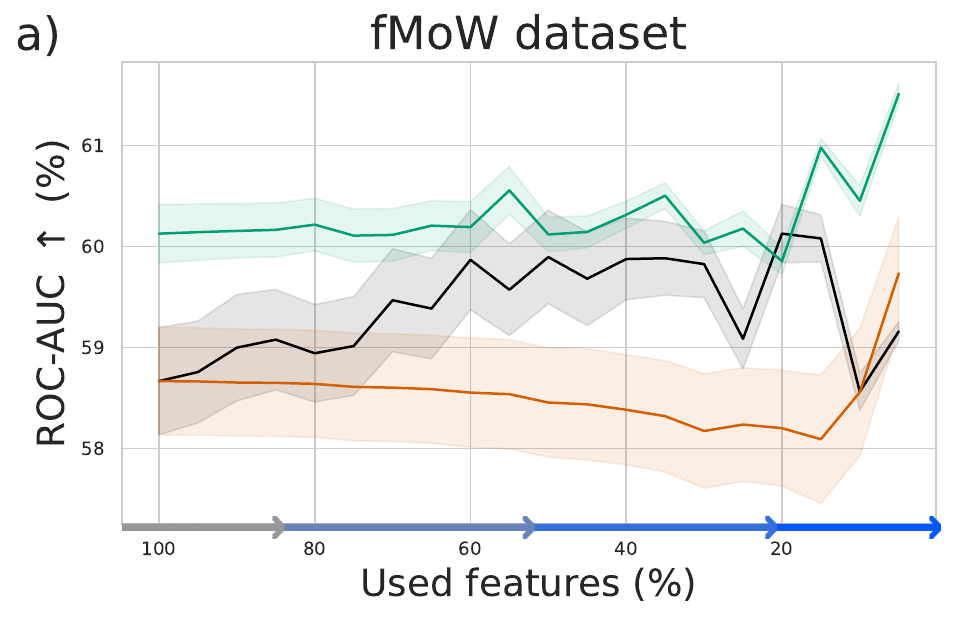}
    \end{subfigure}
    \begin{subfigure}{0.33\textwidth}
        \centering
        \includegraphics[width=\linewidth]{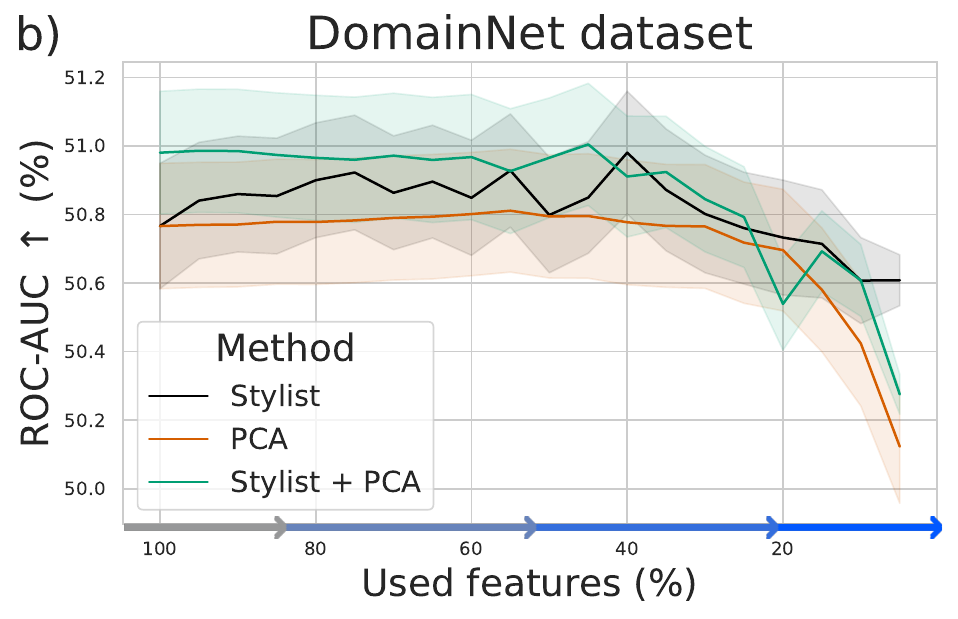}
    \end{subfigure}
    \begin{subfigure}{0.33\textwidth}
        \centering
        \includegraphics[width=\linewidth]{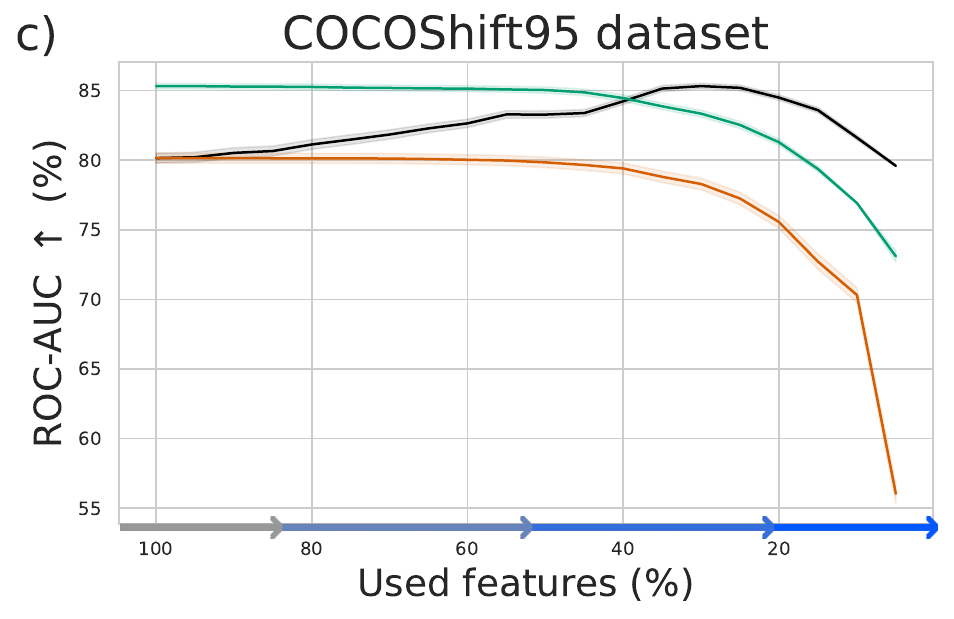}
    \end{subfigure}
     \caption{\textbf{Features Selection vs. Dimensionality Reduction (PCA)}. When comparing Stylist (black) with PCA (orange), we see that Stylist selection works better in all cases. Moreover, when combining the best selection percentage of Stylist with further dimensionality reduction using PCA (green), we observe an improvement (note that the green curve corresponds to different absolute numbers of features).}
    \label{fig:selection_vs_pca}
\end{figure*}

\noindent\textbf{Stylist robustness to dataset spuriousness level} 
    To better understand the real cases, we further analyze the impact of spurious correlations in each step of our approach. We use datasets with various levels of {\it spuriousness} between style and content, in three setups (Fig.~\ref{fig:spuriousness}): {\bf a)} use the same dataset in both algorithm steps; {\bf b)} keep the spuriousity level fixed for Step2 while varying the spuriousity level for Step1 {\bf c)} keep the spuriousity level fixed for Step1 while varying the spuriousity level for Step2.
    %keep the ND training dataset constant, while changing the features ranking one; {\bf c)} keep the features ranking dataset constant, while variate the ND training one. 
    The dataset kept constant in b) and c) is COCOShift\_balanced. We observe that having a higher degree of spuriousness in feature selection (Step 1), leads to better performance for our Stylist method. Nevertheless, in all cases, even in the most degenerated ones (with very high correlations to none), we see an increase after removing the top-ranked environment-biased features.

\noindent\textbf{Feature Selection vs. Dimensionality Reduction} % de ce; ce; cum; rezultate
Classical dimensionality reduction approaches (like \textbf{PCA}) address the idea of reducing space dimensionality while preserving or maximizing the most important information. In PCA, we can assume that a projection into the space of the principal components will produce a robust representation. Although this projection method differs from feature selection methods, as it reprojects features into a new space rather than retaining specific features, we compare it with {\it Stylist}, in Fig.~\ref{fig:selection_vs_pca}, for the robust novelty detection task. Consistently, for all datasets, {\it Stylist} selection performs better. We also combine {\it Stylist} with PCA, by applying an additional dimensionality reduction over the best percentage of features from {\it Stylist}. We observe an improvement in the curves, highlighting the potential of combining the two approaches, proving that the two methods are not only different, but also complementary.
% Features selection is different from dimensionality reduction, in the sense that the first keeps certain existing features, while the second reprojects all of the features in a new space.
% So even though {\it Stylist} is conceptually different from PCA, we compare in Fig.~\ref{fig:selection_vs_pca} the ND OOD performance after selection and after projection.
%Consistently, for all datasets, {\it Stylist} selection performs better. More, in some cases we see an improvement in the curves when we combine the two (on DomainNet and fMoW). For this, we first select the features with our {\it Stylist}, choose the best percentage of features to keep and next we project the rest using PCA.

\subsection{Ablations}
\noindent\textbf{Feature extractors} We show in Tab.~\ref{tab:feature_extractors} that our feature selection method is model-agnostic, improving over 100\% feature usage baseline, over a wide variety of pretrained models, coming from basic supervised classification, multi-modal and contrastive approaches.

\begin{figure*}[t]
    \begin{subfigure}{0.4\textwidth}
        \centering
        \includegraphics[width=\columnwidth]{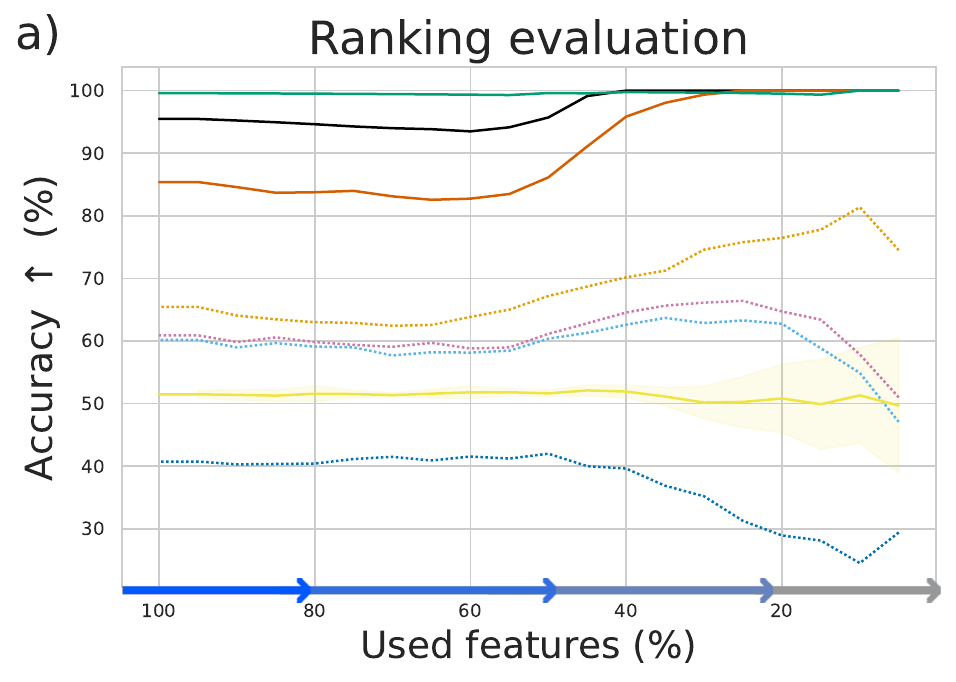}
    \end{subfigure}
     \begin{subfigure}{0.15\textwidth}
     \hfill
        \includegraphics[width=\columnwidth]{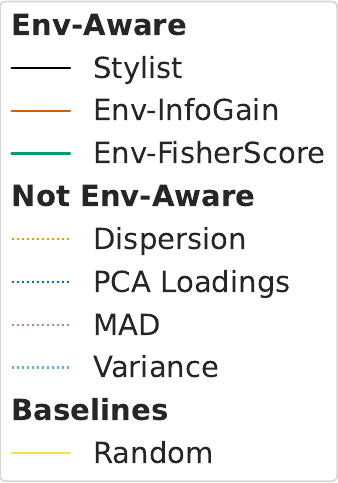}    
        \hfill
    \end{subfigure}
    \begin{subfigure}{0.4\textwidth}
        \centering
        \includegraphics[width=\columnwidth]{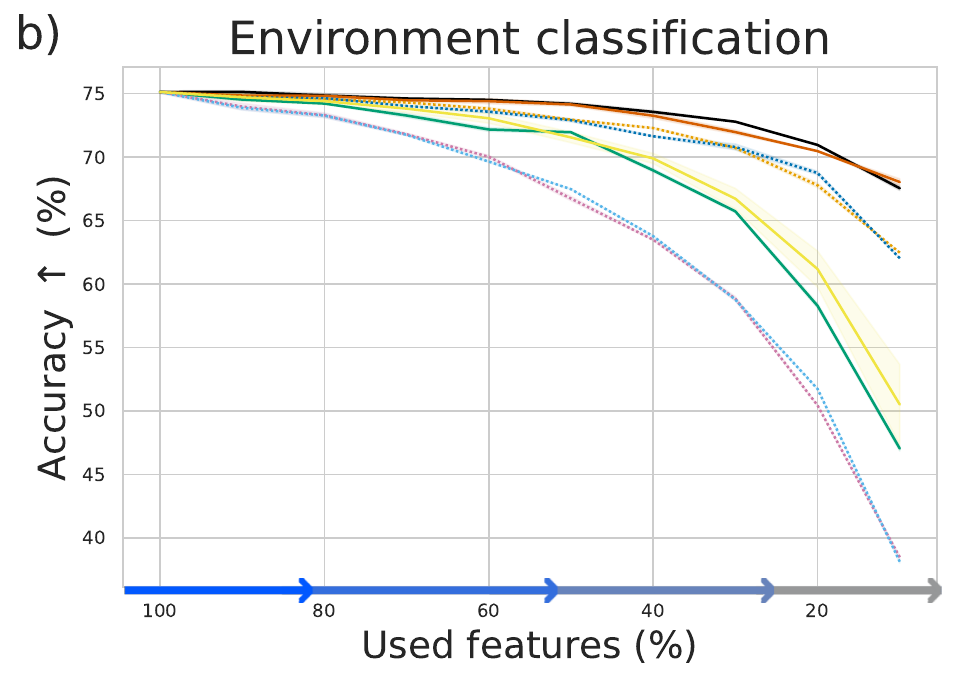}
    \end{subfigure}
     \caption{\textbf{Features’ focus analysis}. {\bf a)} In a controlled experiment, with 50\% of features being content-related and 50\% being style-related, we evaluate how accurate is Stylist in determine which is which. We observe that the top-ranked 40\% features are correctly identified as environment related (100\% accuracy in predicting whether a feature is content or style related). In fact, all {\it env-aware} methods have impressive results, overcoming {\it non env-aware} methods by a large margin. {\bf b)} In a balanced setup, we have also evaluated the ability of our top-ranked environment-biased features to classify the style category of an image. Note that our approach reaches a high accuracy with only 5\% of the top-ranked environment-biased features. This indicates that the identified features are indeed strongly correlated with the style.}
    \label{fig:feature_focus}
\end{figure*}

\noindent\textbf{Stylist distance} % de ce; ce; cum; rezultate
We validate the algorithmic decisions of our proposed {\it Stylist} approach. To compute the per-feature scores, we measure the per-feature distance in distribution (Eq.~\ref{eq:f_distr}), between any two training environments (Eq.~\ref{eq:feature_ranking_1}), and combine those per-pair distances to obtain a more informative ranking, based on all training environments (Eq.~\ref{eq:feature_ranking_2}). The per-pair ranking combinations do not influence the overall performance, while the distance used seems to be dataset-specific (symmetric KL is better on fMoW, while Wasserstein is better on DomainNet and the synthetic COCOShift95). For simplicity, we have used Wasserstein distance with mean over the features per-pair scores in all our experiments. See Appendix~\ref{apx:ablation_distance} for detailed scores.

\noindent\textbf{The percent of selected features} As highlighted in Fig.~\ref{fig:sota_feature_selection}, Stylist consistently improves over the baseline w.r.t. the percent of considered features, proving that the provided feature ranking is relevant for the novelty detection problem. To select an optimal percent of features per setup, we employ a validation step, analyzing either the performance on an ID validation set or the performance on an OOD test set. There is a very small performance variance between those techniques (less than 0.015 for ResNet-18 features, and even 0 for CLIP features), highlighting that the performance is stable. This is an important property of our algorithm, managing to improve Out-Of-Distribution performance, with In-Distribution chosen hyper-parameters.

\noindent\textbf{Class (un)balanced environments} Our method benefits from having the same distribution of classes in all environments. While we assume that the class variations are small, we do not strictly enforce this in our main experiments, as we assume that we do not have access to the class labels. Nevertheless, we can use the class label information to have the same class distribution in all environments, by resampling. Although the class distributions in our datasets vary by a small amount between environments, we performed an experiment where we resampled all datasets to be class-balanced and applied the selection method to the resulting ones. That ROC-AUC stays the same or improves slightly (+0.87\%) compared to the initial unbalanced dataset.

\subsection{A glimpse of interpretability}

Our approach ranks features based on how much they represent the environment's irrelevant factors, and we validate their quality with two experiments.

% Our approach ranks features based on their correlation with the style. To validate the quality of the ranking, we perform two experiments.

\textbf{a)} We investigate if the approach can find the features that focus on the environment factors in the ideal case of disentangled features, where some features exclusively represent style while others exclusively represent content.
For this, we split each image from COCOShift\_balanced train set into two images, one containing only the object (content), while the other only the background (style). We independently extract features from the two images and then concatenate them, thus the first 50\% of the features are content features and the rest are style features.
% \textbf{a)} % 
% We build a synthetic scenario with disentangled features for style and content. 
% We split each image from COCOShift\_balanced train set into two images. One only contains the object (content), and the other only contains the background (style). We extract and concatenate the features for the two images. The first half of the features are unrelated to style, while the second half is unrelated to content (for evaluation purposes we have considered that the first 50\% of features are content features and the rest are style features).
Further, we apply \textit{Stylist} over this combined representation, on COCOShift\_balanced dataset. 
% nu am zis exact cum se calculeaza acuratetea aia
For each percent of features used (from 5\% to 100\%) we compute the accuracy of this selection (first 50\% should be environment features then style features).
In Fig.~\ref{fig:feature_focus} a) we present the results of our experiment. For the first 40\% top-ranked environment-biased features, {\it Stylist} has a perfect accuracy score, with other {\it env-aware} methods (Env-InfoGain and Env-FisherScore) having also impressive scores of 99.1\% and 99.7\%, while the {\it non env-aware} methods performing significantly lower. In this scenario with disentangled features, env-aware methods consistently select as top features, those associated with the style.

\textbf{b)} In a more realistic scenario, with pretrained features that might not be disentangled, we analyze the degree in which the top-scored environment-biased features represent style. For this, we train a classifier to predict the ground-truth style of an image, given the selected features.
% In a real case scenario, where we know nothing about the meaning of each feature, we analyze the capacity of our top-scored environment-biased features to predict the style of an image. 
Starting from our COCOShift dataset, we build a balanced dataset, without spurious correlations, for the task of classifying the style category of an image (1 out of 9). In Fig.~\ref{fig:feature_focus} b) we present the results of our experiment, where we have trained a classifier for each percent of features. We observe that with only a small fraction of the features, we achieve almost the maximum score for predicting the environment, showing that the top-ranked features are, indeed, style predictive. In contrast, when randomly selecting features, the same performance is achieved using significantly more features.

Although the FisherScore selection method works very well in \textbf{a)}, in the perfect feature disentanglement case, in the real-world scenario of \textbf{ b)}, it fails below the random baseline. Intuitively, FisherScore relies on computing full feature space distances when finding neighbours, but those distances are largely affected by the imperfect scenario, where features are intertwined and the feature extractor can contain an unbalanced ratio of style vs. content features. In contrast, {\it Stylist} analyzes distances between individual feature distributions, implicitly balancing the impact of content vs. style if part of the spectrum looks similar because it represents the content part. In this way, {\it Stylist} manages to be more robust in the real-case scenario like in \textbf{b)}.

\section{Related work}
\textbf{Out-Of-Domain (OOD) generalization}:
Machine learning methods proved to have remarkable capabilities, but are still subject to mistakes when dealing with out-of-distribution data~\cite{geirhos2020shortcut, beery2018recognition,hendrycks2021natural, lechner2022all}. 

\textbf{Invariant learning}: To tackle the changing distribution, one possible solution involves learning some invariant mechanisms of the data~\cite{muandet2013domain, peters2016causal, irm}. IRM~\cite{irm} constraints the model such to obtain the same classifier in different environments, while vREx~\cite{vrex} constrains the loss to have low variance across domains. 
% for camera ready or later: andrei shameless self-citation: 
InvariantKNN~\cite{iknn} obtains invariant representations by constraining the features to produce the same KNN predictor across environments.
The work of~\cite{ye2021towards} proves that features with small variations between training environments are important for out-of-distribution generalization. This also gives a formal motivation to our work. In~\cite{wang2022provable_inv_feat_recov} a subspace of invariant features is determined through PCA of class-embeddings. A formalization of invariant learning is proposed in~\cite{cisa} and suggests that depending on the structure of the data, different constraints should be used.  Different from those solutions that require both semantic (content) labels (namely content classes) and environment labels, Stylist needs only environment labels.

%Since the semantic labels are available in our used datasets, we did an experiment that focuses on the complementarity between the methods, not on their differences: We first trained IRM on those semantic labels. Next, we applied Stylist on top of them. Interestingly, we see a $1\%$ increase over using $100\%$ of the features, showing that the two methods are somehow complementary.

\textbf{OOD datasets}: Other existing datasets on OOD~\cite{koh2021wilds} have different limitations when we tried to approach them using Stylist, motivating us to introduce COCOShift, which brings a controllable level of spurious correlations. Waterbirds~\cite{waterbirds} has only two environments, one used for training, and one used for testing. The existence of multiple environments is essential to define what the style consists of, so we need to see at least two training domains to determine the environment-biased features. Background challenge benchmark~\cite{background_challenge} does not provide labels for the style, but only for the content. Furthermore, by construction, we have secondary objects in the samples, for which we don’t have labels, but they come from the same set of classes as the main content. This makes it impossible to have a clear separation between style and content. In MetaShift~\cite{metashift}, the annotations of each image encompass nearly all objects present within it. Usually, there are 3 or more different objects per image, which intersect the content and environment label sets. We only found 2159 clean samples (with no conflicting annotations). For non-intersecting content and environment label sets, there are only 281 samples.

\textbf{Novelty detection}: 
Semantic anomaly detection~\cite{ahmed2020_semantic_anomalies} aims to detect only changes in some high-level semantic factors (\eg object classes) as opposed to low-level cues (such as image artifacts). Methods like the ones in~\cite{tack2020csi, sehwag2021ssd, winkens2020contrastive, sun2022out} use a self-supervised method for anomaly or out-of-distribution detection while the methods in~\cite{li2021cutpaste, zhou2021contrastive, reiss2023mean_shifted} also adapt pretrained extractors using contrastive methods. 
% \andrei{maybe add \cite{vojir2023calibrated} care folosesc tot pretrained representations} 
RedPanda~\cite{cohen2023red_panda} method learns to ignore some irrelevant factors but achieves this using labels of such factors. Still, most works in this space only focus on settings containing only one type of factor, semantic or non-semantic, but not both.

\textbf{Robust novelty detection}: We propose this term for the setting that contains both content and style factors, where the goal is to detect changes in content while being robust to style. This setting is introduced in~\cite{ea_moco} where they show that robustness methods based on multi-environment learning can help anomaly detection. Our work shows that a simple, but efficient method of ranking feature invariance improves performance in the context of \textbf{robust novelty detection}.

% \textbf{Open Set Recognition}: Open Set Recognition (OSR) requires more supervision than Novelty Detection (ND), as during training, OSR has access to the semantic labels of the normal / known data. Meanwhile, ND observes during training the set of normal classes, without having access to their specific content labels. OSR is typically approached in a supervised learning context, while ND methods are often employed in an unsupervised context.

% \paragraph{Limitations}
% * We start with a fix representation
% * features nu sunt foarte disentangled, dar noi le taiem pe alea cel mai probabil sa fie de stil.

\section{Conclusions}
In this work, we first propose {\it Stylist}, a feature selection method that finds features focused more on the environment, which are irrelevant for a pursued task, by emphasizing the distribution distances between environments, at the feature level. Next, we prove that by dropping features for which our algorithm gives a high probability of being environment-biased, we improve the generalization performance of novelty detection in the setup where both style and content distribution shifts. We validate our approach on real-world datasets DomainNet and fMoW as well as our introduced benchmark, COCOShift where we can control the level of spuriousness.
% To validate our approach, we introduce COCOShift, a synthetic benchmark, on which we tested our solution on splits with various levels of spuriousness, alongside other two datasets, DomainNet and fMoW, composed of real sampled data. 

\section{Impact statement}
This paper tackles fundamental research in Machine Learning without any specific application. 
%This paper presents work whose goal is to advance the field of Machine Learning.
As the proposed method is generic, we feel it does not present special or direct ethical or societal negative consequences.
By removing spurious correlations, this approach has the potential for increased fairness and robustness while also being useful for analyzing existing biases in pretrained representations.

\section{Acknowledgments} Funded in part by the EU Horizon project ELIAS (No. 101120237).

%%%%%%%%% REFERENCES
{\small
\bibliographystyle{ieee_fullname}
\bibliography{egbib}

\begin{thebibliography}{10}\itemsep=-1pt

\bibitem{ahmed2020_semantic_anomalies}
Faruk Ahmed and Aaron Courville.
\newblock Detecting semantic anomalies.
\newblock In {\em Proceedings of the AAAI Conference on Artificial Intelligence}, 2020.

\bibitem{knn}
Fabrizio Angiulli and Clara Pizzuti.
\newblock Fast outlier detection in high dimensional spaces.
\newblock In {\em Principles of Data Mining and Knowledge Discovery, {PKDD}}, 2002.

\bibitem{irm}
Martin Arjovsky, L{\'e}on Bottou, Ishaan Gulrajani, and David Lopez-Paz.
\newblock Invariant risk minimization.
\newblock {\em arXiv preprint arXiv:1907.02893}, 2019.

\bibitem{beery2018recognition}
Sara Beery, Grant Van~Horn, and Pietro Perona.
\newblock Recognition in terra incognita.
\newblock In {\em Proceedings of the European conference on computer vision (ECCV)}, 2018.

\bibitem{journals/dss/BhattacharyyaJTW11}
Siddhartha Bhattacharyya, Sanjeev Jha, Kurian~K. Tharakunnel, and J.~Christopher Westland.
\newblock Data mining for credit card fraud: A comparative study.
\newblock {\em Decis. Support Syst.}, 2011.

\bibitem{lof}
Markus~M. Breunig, Hans{-}Peter Kriegel, Raymond~T. Ng, and J{\"{o}}rg Sander.
\newblock {LOF:} identifying density-based local outliers.
\newblock In {\em {SIGMOD} International Conference on Management of Data}, 2000.

\bibitem{Chauhan2015AnomalyDI}
Sucheta Chauhan and Lovekesh Vig.
\newblock Anomaly detection in ecg time signals via deep long short-term memory networks.
\newblock {\em 2015 IEEE International Conference on Data Science and Advanced Analytics (DSAA)}, 2015.

\bibitem{fmow}
Gordon~A. Christie, Neil Fendley, James Wilson, and Ryan Mukherjee.
\newblock Functional map of the world.
\newblock In {\em {IEEE} Conference on Computer Vision and Pattern Recognition, {CVPR}}, 2018.

\bibitem{cohen2023red_panda}
Niv Cohen, Jonathan Kahana, and Yedid Hoshen.
\newblock Red {PANDA}: Disambiguating image anomaly detection by removing nuisance factors.
\newblock In {\em The Eleventh International Conference on Learning Representations}, 2023.

\bibitem{imagenet}
Jia Deng, Wei Dong, Richard Socher, Li{-}Jia Li, Kai Li, and Li Fei{-}Fei.
\newblock Imagenet: {A} large-scale hierarchical image database.
\newblock In {\em {IEEE} Computer Society Conference on Computer Vision and Pattern Recognition {(CVPR} 2009)}, 2009.

\bibitem{anoshift}
Marius Dragoi, Elena Burceanu, Emanuela Haller, Andrei Manolache, and Florin Brad.
\newblock Anoshift: A distribution shift benchmark for unsupervised anomaly detection.
\newblock {\em Advances in Neural Information Processing Systems}, 35:32854--32867, 2022.

\bibitem{geirhos2020shortcut}
Robert Geirhos, J{\"o}rn-Henrik Jacobsen, Claudio Michaelis, Richard Zemel, Wieland Brendel, Matthias Bethge, and Felix~A Wichmann.
\newblock Shortcut learning in deep neural networks.
\newblock {\em Nature Machine Intelligence}, 2(11), 2020.

\bibitem{fisher}
Quanquan Gu, Zhenhui Li, and Jiawei Han.
\newblock Generalized fisher score for feature selection.
\newblock In {\em {UAI} Proceedings of the Twenty-Seventh Conference on Uncertainty in Artificial Intelligence}, 2011.

\bibitem{resnet}
Kaiming He, Xiangyu Zhang, Shaoqing Ren, and Jian Sun.
\newblock Deep residual learning for image recognition.
\newblock In {\em 2016 {IEEE} Conference on Computer Vision and Pattern Recognition {CVPR}}, 2016.

\bibitem{hendrycks2021natural}
Dan Hendrycks, Kevin Zhao, Steven Basart, Jacob Steinhardt, and Dawn Song.
\newblock Natural adversarial examples.
\newblock In {\em Proceedings of the IEEE/CVF Conference on Computer Vision and Pattern Recognition}, pages 15262--15271, 2021.

\bibitem{align}
Chao Jia, Yinfei Yang, Ye Xia, Yi{-}Ting Chen, Zarana Parekh, Hieu Pham, Quoc~V. Le, Yun{-}Hsuan Sung, Zhen Li, and Tom Duerig.
\newblock Scaling up visual and vision-language representation learning with noisy text supervision.
\newblock In {\em Proceedings of the 38th International Conference on Machine Learning, {ICML}}, 2021.

\bibitem{supcon}
Prannay Khosla, Piotr Teterwak, Chen Wang, Aaron Sarna, Yonglong Tian, Phillip Isola, Aaron Maschinot, Ce Liu, and Dilip Krishnan.
\newblock Supervised contrastive learning.
\newblock In {\em Advances in Neural Information Processing Systems, NeurIPS}, 2020.

\bibitem{DFR}
Polina Kirichenko, Pavel Izmailov, and Andrew~Gordon Wilson.
\newblock Last layer re-training is sufficient for robustness to spurious correlations.
\newblock In {\em The Eleventh International Conference on Learning Representations}, 2023.

\bibitem{koh2021wilds}
Pang~Wei Koh, Shiori Sagawa, Henrik Marklund, Sang~Michael Xie, Marvin Zhang, Akshay Balsubramani, Weihua Hu, Michihiro Yasunaga, Richard~Lanas Phillips, Irena Gao, et~al.
\newblock Wilds: A benchmark of in-the-wild distribution shifts.
\newblock In {\em International Conference on Machine Learning}. PMLR, 2021.

\bibitem{infogain}
Alexander Kraskov, Harald Stoegbauer, and Peter Grassberger.
\newblock Estimating mutual information.
\newblock In {\em Phys. Rev}, 2004.

\bibitem{vrex}
David Krueger, Ethan Caballero, J{\"{o}}rn{-}Henrik Jacobsen, Amy Zhang, Jonathan Binas, Dinghuai Zhang, R{\'{e}}mi~Le Priol, and Aaron~C. Courville.
\newblock Out-of-distribution generalization via risk extrapolation (rex).
\newblock In {\em Proceedings of the 38th International Conference on Machine Learning, {ICML}}, 2021.

\bibitem{lechner2022all}
Mathias Lechner, Ramin Hasani, Alexander Amini, Tsun-Hsuan Wang, Thomas~A Henzinger, and Daniela Rus.
\newblock Are all vision models created equal? a study of the open-loop to closed-loop causality gap.
\newblock {\em arXiv preprint arXiv:2210.04303}, 2022.

\bibitem{kloft2023zero}
Aodong Li, Chen Qiu, Marius Kloft, Padhraic Smyth, Maja Rudolph, and Stephan Mandt.
\newblock Zero-shot anomaly detection without foundation models.
\newblock {\em arXiv preprint arXiv:2302.07849}, 2023.

\bibitem{li2021cutpaste}
Chun-Liang Li, Kihyuk Sohn, Jinsung Yoon, and Tomas Pfister.
\newblock Cutpaste: Self-supervised learning for anomaly detection and localization.
\newblock In {\em Proceedings of the IEEE/CVF conference on computer vision and pattern recognition}, 2021.

\bibitem{blip2}
Junnan Li, Dongxu Li, Silvio Savarese, and Steven C.~H. Hoi.
\newblock {BLIP-2:} bootstrapping language-image pre-training with frozen image encoders and large language models.
\newblock In {\em International Conference on Machine Learning, {ICML}}, 2023.

\bibitem{metashift}
Weixin Liang and James Zou.
\newblock Metashift: A dataset of datasets for evaluating contextual distribution shifts and training conflicts.
\newblock In {\em ICLR}, 2022.

\bibitem{COCO}
Tsung{-}Yi Lin, Michael Maire, Serge~J. Belongie, Lubomir~D. Bourdev, Ross~B. Girshick, James Hays, Pietro Perona, Deva Ramanan, Piotr Doll{\'{a}}r, and C.~Lawrence Zitnick.
\newblock Microsoft {COCO:} common objects in context.
\newblock {\em CoRR}, abs/1405.0312, 2014.

\bibitem{muandet2013domain}
Krikamol Muandet, David Balduzzi, and Bernhard Sch{\"o}lkopf.
\newblock Domain generalization via invariant feature representation.
\newblock In {\em International Conference on Machine Learning}. PMLR, 2013.

\bibitem{iknn}
Andrei~Liviu Nicolicioiu, Jerry Huang, Dhanya Sridhar, and Aaron Courville.
\newblock Do as your neighbors: Invariant learning through non-parametric neighbourhood matching.
\newblock In {\em ICML Workshop on Spurious Correlations, Invariance and Stability (SCIS)}. 2023.

\bibitem{DomainNet}
Xingchao Peng, Qinxun Bai, Xide Xia, Zijun Huang, Kate Saenko, and Bo Wang.
\newblock Moment matching for multi-source domain adaptation.
\newblock In {\em Proceedings of the IEEE/CVF International Conference on Computer Vision (ICCV)}, October 2019.

\bibitem{peters2016causal}
Jonas Peters, Peter B{\"u}hlmann, and Nicolai Meinshausen.
\newblock Causal inference by using invariant prediction: identification and confidence intervals.
\newblock {\em Journal of the Royal Statistical Society: Series B (Statistical Methodology)}, 2016.

\bibitem{pimentel2014review}
Marco~AF Pimentel, David~A Clifton, Lei Clifton, and Lionel Tarassenko.
\newblock A review of novelty detection.
\newblock {\em Signal processing}, 2014.

\bibitem{clip}
Alec Radford, Jong~Wook Kim, Chris Hallacy, Aditya Ramesh, Gabriel Goh, Sandhini Agarwal, Girish Sastry, Amanda Askell, Pamela Mishkin, Jack Clark, Gretchen Krueger, and Ilya Sutskever.
\newblock Learning transferable visual models from natural language supervision.
\newblock In {\em Proceedings of the 38th International Conference on Machine Learning, {ICML}}, 2021.

\bibitem{reiss2023mean_shifted}
Tal Reiss and Yedid Hoshen.
\newblock Mean-shifted contrastive loss for anomaly detection.
\newblock In {\em Proceedings of the AAAI Conference on Artificial Intelligence}, 2023.

\bibitem{ruff2021unifying}
Lukas Ruff, Jacob~R Kauffmann, Robert~A Vandermeulen, Gr{\'e}goire Montavon, Wojciech Samek, Marius Kloft, Thomas~G Dietterich, and Klaus-Robert M{\"u}ller.
\newblock A unifying review of deep and shallow anomaly detection.
\newblock {\em Proceedings of the IEEE}, 2021.

\bibitem{waterbirds}
Shiori Sagawa, Pang~Wei Koh, Tatsunori~B. Hashimoto, and Percy Liang.
\newblock Distributionally robust neural networks for group shifts: On the importance of regularization for worst-case generalization.
\newblock {\em ArXiv}, 2019.

\bibitem{DRO}
Shiori Sagawa*, Pang~Wei Koh*, Tatsunori~B. Hashimoto, and Percy Liang.
\newblock Distributionally robust neural networks.
\newblock In {\em International Conference on Learning Representations}, 2020.

\bibitem{OODsurvey}
Mohammadreza Salehi, Hossein Mirzaei, Dan Hendrycks, Yixuan Li, Mohammad~Hossein Rohban, and Mohammad Sabokrou.
\newblock A unified survey on anomaly, novelty, open-set, and out of-distribution detection: Solutions and future challenges.
\newblock {\em Transactions on Machine Learning Research}, 2022.

\bibitem{ocsvm}
Bernhard Sch{\"{o}}lkopf, Robert~C. Williamson, Alexander~J. Smola, John Shawe{-}Taylor, and John~C. Platt.
\newblock Support vector method for novelty detection.
\newblock In {\em Advances in Neural Information Processing Systems, {NIPS}}, 1999.

\bibitem{sehwag2021ssd}
Vikash Sehwag, Mung Chiang, and Prateek Mittal.
\newblock Ssd: A unified framework for self-supervised outlier detection.
\newblock {\em arXiv preprint arXiv:2103.12051}, 2021.

\bibitem{ea_moco}
Stefan Smeu, Elena Burceanu, Andrei~Liviu Nicolicioiu, and Emanuela Haller.
\newblock Env-aware anomaly detection: Ignore style changes, stay true to content!
\newblock {\em NeurIPSW on Distribution Shifts}, 2022.

\bibitem{fast-knn-ood}
Yiyou Sun, Yifei Ming, Xiaojin Zhu, and Yixuan Li.
\newblock Out-of-distribution detection with deep nearest neighbors.
\newblock In {\em International Conference on Machine Learning, {ICML}}, 2022.

\bibitem{sun2022out}
Yiyou Sun, Yifei Ming, Xiaojin Zhu, and Yixuan Li.
\newblock Out-of-distribution detection with deep nearest neighbors.
\newblock In {\em International Conference on Machine Learning}. PMLR, 2022.

\bibitem{tack2020csi}
Jihoon Tack, Sangwoo Mo, Jongheon Jeong, and Jinwoo Shin.
\newblock Csi: Novelty detection via contrastive learning on distributionally shifted instances.
\newblock {\em Advances in neural information processing systems}, 2020.

\bibitem{wang2022provable_inv_feat_recov}
Haoxiang Wang, Haozhe Si, Bo Li, and Han Zhao.
\newblock Provable domain generalization via invariant-feature subspace recovery.
\newblock In {\em International Conference on Machine Learning}. PMLR, 2022.

\bibitem{cisa}
Zihao Wang and Victor Veitch.
\newblock A unified causal view of domain invariant representation learning.
\newblock In {\em ICML 2022: Workshop on Spurious Correlations, Invariance and Stability}, 2022.

\bibitem{winkens2020contrastive}
Jim Winkens, Rudy Bunel, Abhijit~Guha Roy, Robert Stanforth, Vivek Natarajan, Joseph~R Ledsam, Patricia MacWilliams, Pushmeet Kohli, Alan Karthikesalingam, Simon Kohl, et~al.
\newblock Contrastive training for improved out-of-distribution detection.
\newblock {\em arXiv preprint arXiv:2007.05566}, 2020.

\bibitem{background_challenge}
Kai~Yuanqing Xiao, Logan Engstrom, Andrew Ilyas, and Aleksander Madry.
\newblock Noise or signal: The role of image backgrounds in object recognition.
\newblock In {\em {ICLR}}, 2021.

\bibitem{yang2021generalized}
Jingkang Yang, Kaiyang Zhou, Yixuan Li, and Ziwei Liu.
\newblock Generalized out-of-distribution detection: A survey.
\newblock {\em arXiv preprint arXiv:2110.11334}, 2021.

\bibitem{ye2021towards}
Haotian Ye, Chuanlong Xie, Tianle Cai, Ruichen Li, Zhenguo Li, and Liwei Wang.
\newblock Towards a theoretical framework of out-of-distribution generalization.
\newblock In A. Beygelzimer, Y. Dauphin, P. Liang, and J.~Wortman Vaughan, editors, {\em Advances in Neural Information Processing Systems}, 2021.

\bibitem{Places365}
Bolei Zhou, Agata Lapedriza, Aditya Khosla, Aude Oliva, and Antonio Torralba.
\newblock Places: A 10 million image database for scene recognition.
\newblock {\em IEEE Transactions on Pattern Analysis and Machine Intelligence}, 2017.

\bibitem{zhou2022domain}
Kaiyang Zhou, Ziwei Liu, Yu Qiao, Tao Xiang, and Chen~Change Loy.
\newblock Domain generalization: A survey.
\newblock {\em IEEE Transactions on Pattern Analysis and Machine Intelligence}, 2022.

\bibitem{zhou2021contrastive}
Wenxuan Zhou, Fangyu Liu, and Muhao Chen.
\newblock Contrastive out-of-distribution detection for pretrained transformers.
\newblock {\em arXiv preprint arXiv:2104.08812}, 2021.

\end{thebibliography}
}

%%%%%%%%%%%%%%%%%%%%%%%%%%%%%%%%%%%%%%%%%%%%%%%%%%%%%%%%%%%%%%%%%%%%%%%%%%%%%%%
%%%%%%%%%%%%%%%%%%%%%%%%%%%%%%%%%%%%%%%%%%%%%%%%%%%%%%%%%%%%%%%%%%%%%%%%%%%%%%%
% APPENDIX
%%%%%%%%%%%%%%%%%%%%%%%%%%%%%%%%%%%%%%%%%%%%%%%%%%%%%%%%%%%%%%%%%%%%%%%%%%%%%%%
%%%%%%%%%%%%%%%%%%%%%%%%%%%%%%%%%%%%%%%%%%%%%%%%%%%%%%%%%%%%%%%%%%%%%%%%%%%%%%%
\newpage
\appendix
\onecolumn

\section{Feature selection methods - detailed}
\label{apx:selection_algo_ablation}

We show in Tab.~\ref{tab:selection_algo_ablation} individual results for multiple feature selection algorithms, grouped into env-aware ones and algorithms that are not env-aware. Please note that we adapt basic algorithms for feature selection to make them env-aware. 

Considered feature selection methods:
\begin{itemize}
    \item env-aware
        \begin{itemize}
            \item InfoGain~\cite{infogain}: We adapt the method to the env-ware setup. We compute the mutual information between each feature and the style labels. High scores indicate a higher dependency between feature and style labels $\rightarrow$ environment-biased feature.
            \item FisherScore~\cite{fisher}: We adapt the method to the env-aware setup. We rank the features based on their relevance for the classification of style categories. 
        \end{itemize}
    \item non environment-aware
    \begin{itemize}
        \item MAD: For one feature, it computes the average of absolute differences between each sample value and the mean value. High MAD values indicate high discriminatory power.
        \item Dispersion: This is computed as the ratio of arithmetic and geometric means. High dispersion implies a higher discriminatory power. 
        \item Variance: Generally, you can use variance to discard zero variance features as being completely uninformative. We have ranked the features based on their variance, considering high-variance features as being more informative. 
        \item PCA Loadings: We compute the contribution of each feature to the set of principal components identified by PCA. 
    \end{itemize}
\end{itemize}

\section{Different shift robustness}
\label{apx:sota_sub_population}

\begin{figure*}[t!]
    \begin{subfigure}{0.32\textwidth}
        \centering
        \includegraphics[width=\linewidth]{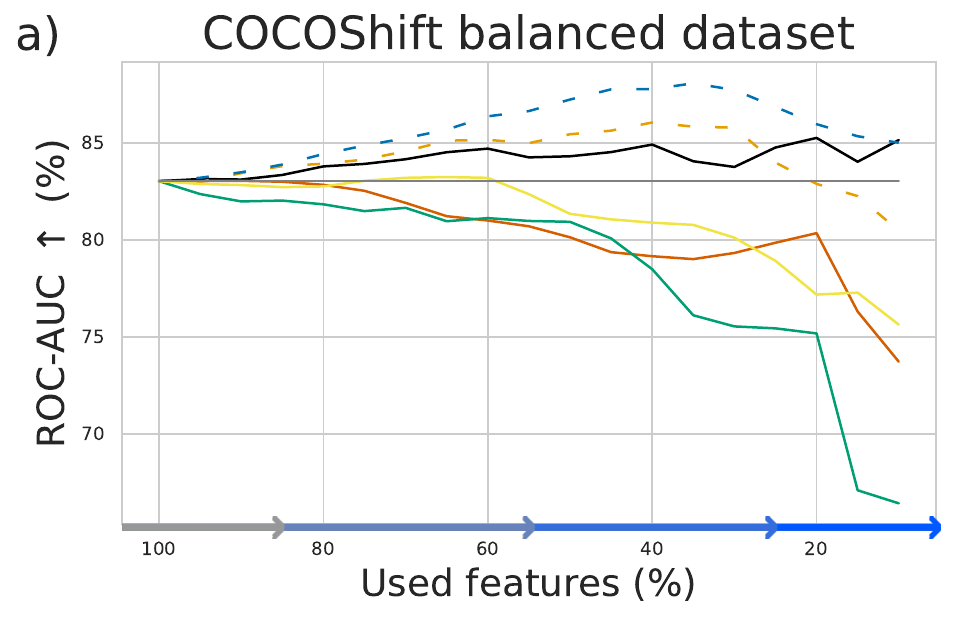}
    \end{subfigure}
    \begin{subfigure}{0.32\textwidth}
        \centering
        \includegraphics[width=\linewidth]{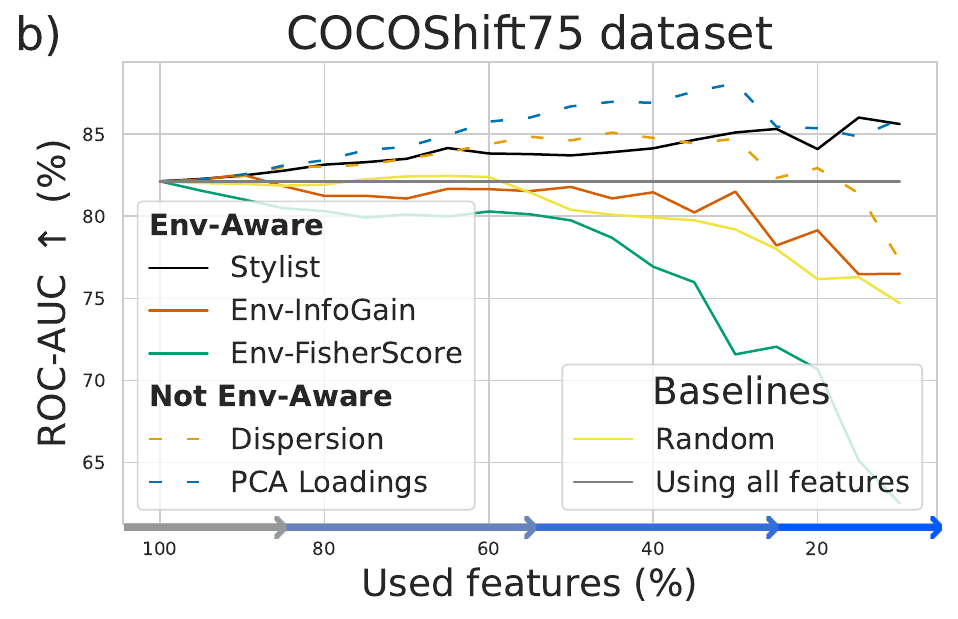}
    \end{subfigure}
    \begin{subfigure}{0.32\textwidth}
        \centering
        \includegraphics[width=\linewidth]{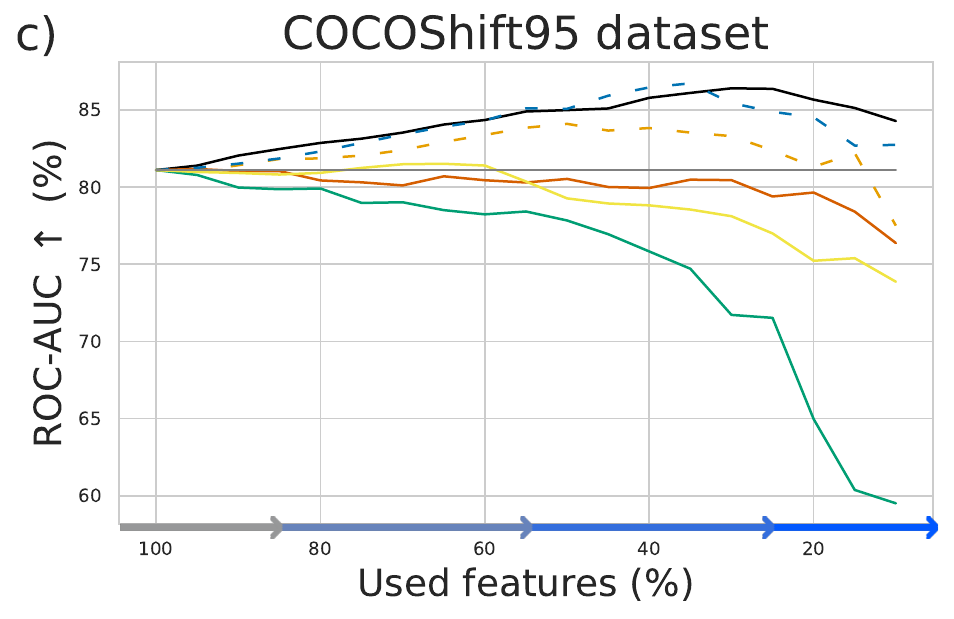}
    \end{subfigure}
    \caption{Feature selection algorithms. The reported ROC-AUC performance is for a testing dataset coming from the same (ID) distribution with a), showing that similar observations remain for sub-population OOD shifts. TODO: a) corresponds to ID setting while c) corresponds to a strong OOD setting}
    \label{fig:sota_feature_selection_sub_pop}
\end{figure*}

We analyzed in Fig.~\ref{fig:sota_feature_selection_sub_pop} what is the impact of {\it Stylist} when we consider 
{\it Sub-Population shifts}, in extension to {\it Covariate shifts}, presented in the main paper. The testing dataset in this case is balanced, ID with the first plot (with no correlation between style and content). Our method manages to improve its performance when compared with other feature selectors, as training and testing become more OOD. The observations are similar in both kinds of shifts.

\begin{table}
    \centering
    % \captionsetup{\textwidth}
    \caption{Feature selection methods.}
    \setlength{\tabcolsep}{6pt}
    \begin{tabular}{ll ccc ccc ccc}
    \toprule
    &\parbox[t]{0mm}{\multirow{5}{*}{\bf \shortstack{Selection\\Method}}}& \multicolumn{3}{c}{\textbf{COCOShift\_balanced}} & \multicolumn{3}{c}{\textbf{COCOShift75}} & \multicolumn{3}{c}{\textbf{COCOShift95}} \\
    \cmidrule(lr){3-5}
    \cmidrule(lr){6-8}
    \cmidrule(lr){9-11}
    && \multicolumn{2}{c}{ROC-AUC $\uparrow$} & \multirow{2}{*}{\shortstack{\% \\selected\\feat.}} & \multicolumn{2}{c}{ROC-AUC $\uparrow$ } & \multirow{2}{*}{\shortstack{\%\\selected\\feat.}} & \multicolumn{2}{c}{ROC-AUC $\uparrow$ } & \multirow{2}{*}{\shortstack{\%\\selected\\feat.}}\\
    \cmidrule(lr){3-4}
    \cmidrule(lr){6-7}
    \cmidrule(lr){9-10}
    && \shortstack{all\\feat.} & \shortstack{\textbf{kept}\\ feat.} & &  \shortstack{all\\feat.} & \shortstack{\textbf{kept}\\ feat.} & & \shortstack{all\\feat.} & \shortstack{\textbf{kept}\\ feat.}\\ 
    \toprule
    % \parbox[t]{2mm}{\multirow{5}{*}{\rotatebox[origin=c]{90}{\textbf{Env-Aware}}}} 
    & \bf Stylist (ours) & 80.9 & 83.5 \footnotesize{\textcolor{ForestGreen}{(+2.6)}} & 40 & 80.4 & \bf 84.7 \footnotesize{\textcolor{ForestGreen}{(+4.3)}} & 10 & 79.8 & \bf 85.1 \footnotesize{\textcolor{ForestGreen}{(+5.3)}} & 30 \\
    &InfoGain & 80.9 & 81.0 \footnotesize{\textcolor{ForestGreen}{(+0.0)}} & 95 & 80.4 & 80.7 \footnotesize{\textcolor{ForestGreen}{(+0.3)}} & 90 & 79.8 & 79.9 \footnotesize{\textcolor{ForestGreen}{(+0.1)}} & 90 \\
    &FisherScore & 80.9 & 80.9 \footnotesize{\textcolor{ForestGreen}{(+0.0)}} & 100 & 80.4 & 80.4 \footnotesize{\textcolor{ForestGreen}{(+0.0)}} & 100 &79.8 & 79.8 \footnotesize{\textcolor{ForestGreen}{(+0.0)}} & 100   \\
    % \cmidrule{2}{2-2}
    % \midrule
    % \midrule
    % \parbox[t]{2mm}{\multirow{6}{*}{\rotatebox[origin=c]{90}{\textbf{Not Env-Aware}}}} 
    & MAD&80.9 & 80.9 \footnotesize{\textcolor{ForestGreen}{(+0.0)}} & 100 &80.4 & 80.4 \footnotesize{\textcolor{ForestGreen}{(+0.0)}} & 100 &79.8 & 79.8 \footnotesize{\textcolor{ForestGreen}{(+0.0)}} & 100 \\
    
    & Dispersion& 80.9 & 84.1 \footnotesize{\textcolor{ForestGreen}{(+3.2)}} & 35 &80.4 & 83.0 \footnotesize{\textcolor{ForestGreen}{(+2.6)}} & 45 &79.8 & 82.2 \footnotesize{\textcolor{ForestGreen}{(+2.4)}} & 50 \\
    & Variance& 80.9 & 80.9 \footnotesize{\textcolor{ForestGreen}{(+0.0)}} & 100 &80.4 & 80.4 \footnotesize{\textcolor{ForestGreen}{(+0.0)}} & 100 &79.8 & 79.8 \footnotesize{\textcolor{ForestGreen}{(+0.0)}} & 100 \\
    & PCA Loadings & 80.9 & \bf 84.6 \footnotesize{\textcolor{ForestGreen}{(+3.7)}} & 35 &80.4 & \bf 84.7 \footnotesize{\textcolor{ForestGreen}{(+4.3)}} & 30 &79.8 & 83.7 \footnotesize{\textcolor{ForestGreen}{(+3.9)}} & 35 \\
    % \cmidrule{lr}{1-2}
    % \midrule
    \\
    \end{tabular}
    \label{tab:selection_algo_ablation}
\end{table}

\section{Stylist ablation distances}
\label{apx:ablation_distance}
\begin{figure*}[t]
    \begin{subfigure}{0.33\textwidth}
        \centering
        \includegraphics[width=\linewidth]{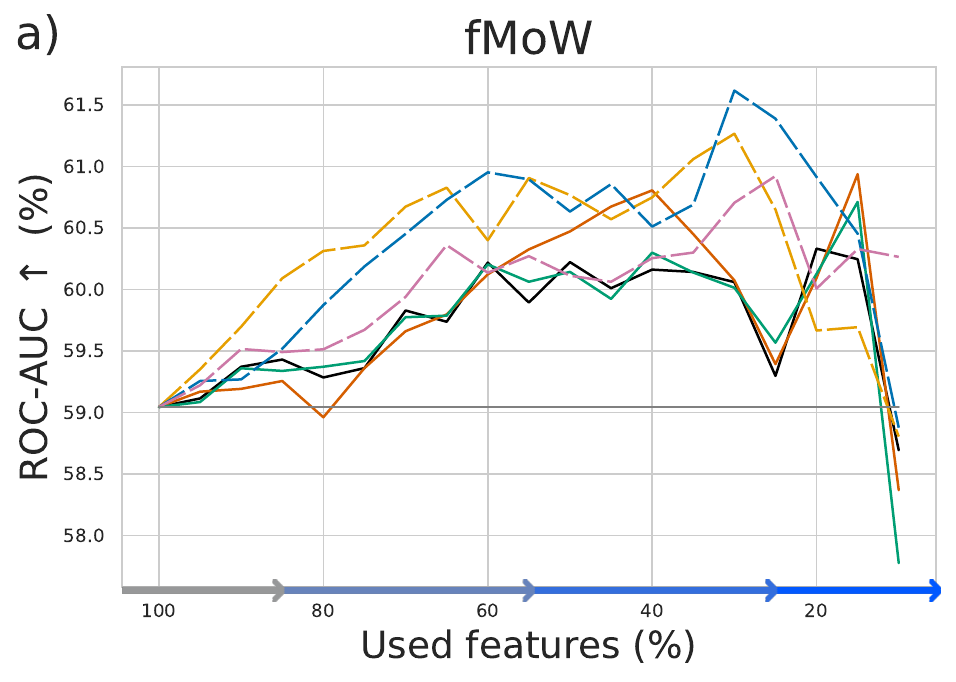}
    \end{subfigure}
    \begin{subfigure}{0.33\textwidth}
        \centering
        \includegraphics[width=\linewidth]{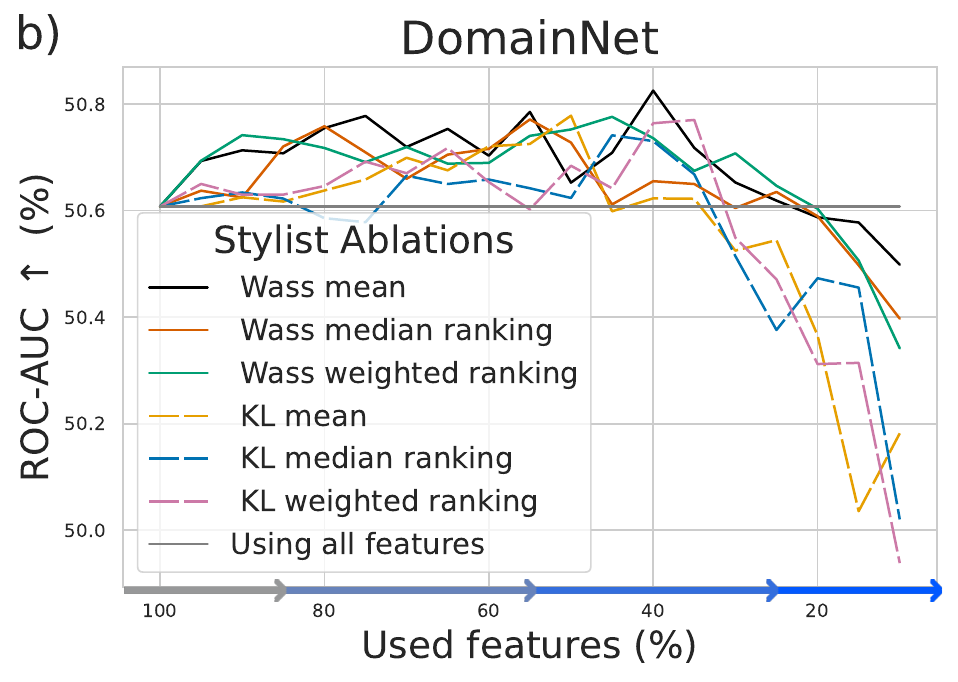}
    \end{subfigure}
    \begin{subfigure}{0.33\textwidth}
        \centering
        \includegraphics[width=\linewidth]{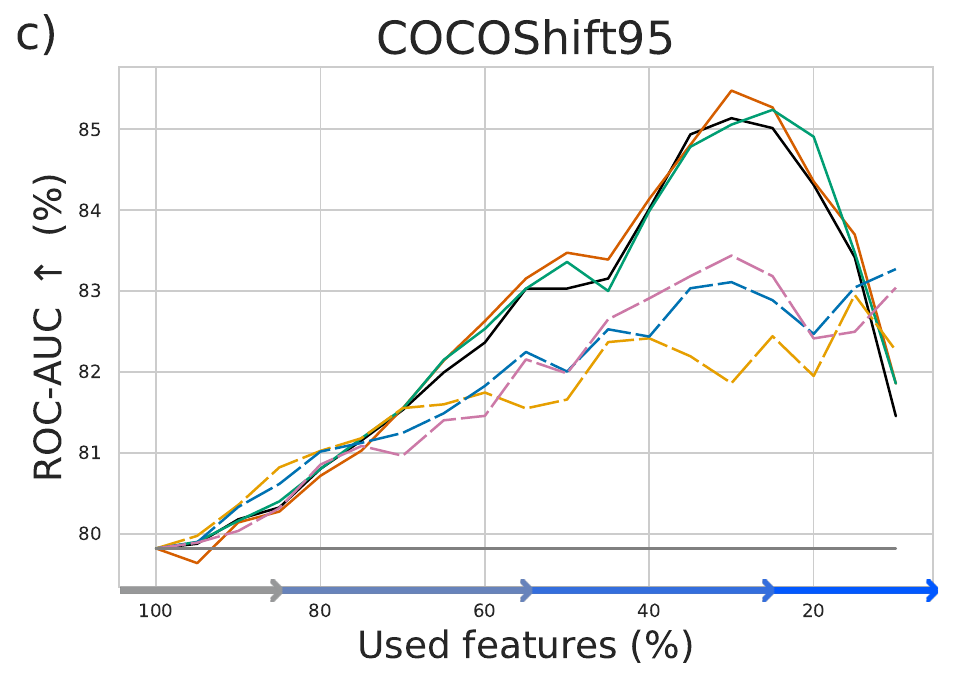}
    \end{subfigure}
    \caption{\textbf{Stylist ablations.} We vary both the distance metrics (based on Wasserstein or symmetric KL) and the ranking combination approaches for pairs of environments (mean, median ranking, weighted ranking). Notice how usually Wasserstein distances perform better (except for a) fMoW). Also, see how the ranking combination does not have a high influence on the result.}
    \label{fig:ablation_kl_wass_distances}
\end{figure*}

In Fig.~\ref{fig:ablation_kl_wass_distances} and Tab.~\ref{tab:ablation_distance} we show results when using symmetric KL or Wasserstein to measure per feature distribution distances between any two training environments. We combine the score or ranking obtained, over all pairs of training envs, using mean, median, median ranking or weighted mean ranking.

\begin{table}[t]
    \centering
    \caption{Ablation distance and ranking combining. We notice here that all variants of distances and feature ranking combinations manage to improve. Depending on the dataset, there are different chooses to make as hyper-parameters. We use for Stylist the mean of Wasserstein distances over all training pairs.}
    \setlength{\tabcolsep}{4pt}
    \begin{tabular}{ll ccc ccc ccc}
    \toprule
    \parbox[t]{0mm}{\multirow{5}{*}{\rotatebox[origin=c]{90}{\bf Distance}}}&\parbox[t]{0mm}{\multirow{5}{*}{\bf Method}}& \multicolumn{3}{c}{\textbf{fMoW}} & \multicolumn{3}{c}{\textbf{DomainNet}} & \multicolumn{3}{c}{\textbf{COCOShift95}} \\
    \cmidrule(lr){3-5}
    \cmidrule(lr){6-8}
    \cmidrule(lr){9-11}
    && \multicolumn{2}{c}{ROC-AUC $\uparrow$} & \multirow{2}{*}{\shortstack{\% \\selected\\feat.}} & \multicolumn{2}{c}{ROC-AUC $\uparrow$ } & \multirow{2}{*}{\shortstack{\%\\selected\\feat.}} & \multicolumn{2}{c}{ROC-AUC $\uparrow$ } & \multirow{2}{*}{\shortstack{\%\\selected\\feat.}}\\
    \cmidrule(lr){3-4}
    \cmidrule(lr){6-7}
    \cmidrule(lr){9-10}
    && \shortstack{all\\feat.} & \shortstack{\textbf{Stylist}\\ feat.} & &  \shortstack{all\\feat.} & \shortstack{\textbf{Stylist}\\ feat.} & & \shortstack{all\\feat.} & \shortstack{\textbf{Stylist}\\ feat.}\\ 
    \toprule
    \multirow{6}{*}{\rotatebox[origin=c]{90}{\bf Wasserstein}}
    \\
    &\bf mean & 59.0 & 60.3 \footnotesize{\textcolor{ForestGreen}{(+1.3)}} & 20 &50.6 & 50.8 \footnotesize{\textcolor{ForestGreen}{(+0.2)}} & 40 &79.8 & 85.1 \footnotesize{\textcolor{ForestGreen}{(+5.3)}} & 30  \\
    &median & 59.0 & 60.6 \footnotesize{\textcolor{ForestGreen}{(+1.5)}} & 15 &50.6 & 50.8 \footnotesize{\textcolor{ForestGreen}{(+0.2)}} & 75 & 79.8 & 84.2 \footnotesize{\textcolor{ForestGreen}{(+4.4)}} & 20  \\
    &median ranking & 59.0 & 60.9 \footnotesize{\textcolor{ForestGreen}{(+1.9)}} & 15 &50.6 & 50.8 \footnotesize{\textcolor{ForestGreen}{(+0.2)}} & 55 & 79.8 & 85.5 \footnotesize{\textcolor{ForestGreen}{(+5.7)}} & 30  \\
    & \shortstack{weighted mean ranking}
    & 59.0 & 60.7 \footnotesize{\textcolor{ForestGreen}{(+1.7)}} & 15 &
    50.6 & 50.8 \footnotesize{\textcolor{ForestGreen}{(+0.2)}} & 45 &
    79.8 & 85.2 \footnotesize{\textcolor{ForestGreen}{(+5.4)}} & 25  \\
    \\
    \midrule
    \multirow{4}{*}{\rotatebox[origin=c]{90}{\bf KL symmetric}} \\
    &mean & 59.0 & 61.3 \footnotesize{\textcolor{ForestGreen}{(+2.2)}} & 30 &50.6 & 50.8 \footnotesize{\textcolor{ForestGreen}{(+0.2)}} & 50 & 79.8 & 83.0 \footnotesize{\textcolor{ForestGreen}{(+3.1)}} & 15\\
    &median & 59.0 & 60.7 \footnotesize{\textcolor{ForestGreen}{(+1.6)}} & 50 &50.6 & 50.8 \footnotesize{\textcolor{ForestGreen}{(+0.2)}} & 45 &79.8 & 83.7 \footnotesize{\textcolor{ForestGreen}{(+3.8)}} & 25 \\
     &median ranking  &59.0 &  61.6 \footnotesize{\textcolor{ForestGreen}{(+2.6)}} & 30 & 50.6 & 50.7 \footnotesize{\textcolor{ForestGreen}{(+0.1)}} & 45 & 79.8 & 83.3 \footnotesize{\textcolor{ForestGreen}{(+3.5)}} & 10\\
   &\shortstack{weighted mean ranking} & 59.0 & 60.9 \footnotesize{\textcolor{ForestGreen}{(+1.9)}} & 25 &50.6 & 50.8 \footnotesize{\textcolor{ForestGreen}{(+0.2)}} & 35 & 79.8 & 83.4 \footnotesize{\textcolor{ForestGreen}{(+3.6)}} & 30 \\
   \\
    \end{tabular}
    \label{tab:ablation_distance}
\end{table}

% \begin{table}[t!]
%     \caption{Selection vs projection}
%     \begin{center}
%     \begin{tabular}{l ll ll ll}
%     \multicolumn{1}{c}{\bf Method}     &\multicolumn{6}{c}{\bf ROC-AUC \% $\uparrow$ (improvement \% $\uparrow$)}\\ 
%      &\multicolumn{2}{c}{\bf fMoW} &\multicolumn{2}{c}{\bf DomainNet} &\multicolumn{2}{c}{\bf COCOShift75}\\
%      &\multicolumn{1}{c}{Best} &\multicolumn{1}{c}{Percent} &\multicolumn{1}{c}{Best} &\multicolumn{1}{c}{Percent} &\multicolumn{1}{c}{Best} &\multicolumn{1}{c}{Percent}\\
%     \hline
%     PCA & 60.1 (+1.1) & 5 & 50.7  & 55 & 80.4  & 100 \\
%     Stylist &  60.3 (+1.3) & 20 &51.5 (+0.9) & 1 &84.7 (+4.3) & 10\\
%     Stylist + PCA \\
%     \end{tabular}
%     \end{center}
%     \label{tab:selection_vs_pca}
% \end{table}

\section{Qualitative analysis}
\label{appendix:qualitative}
\begin{figure}
    \centering

    \begin{subfigure}{0.38\textwidth}
        \centering
        \includegraphics[width=\textwidth]{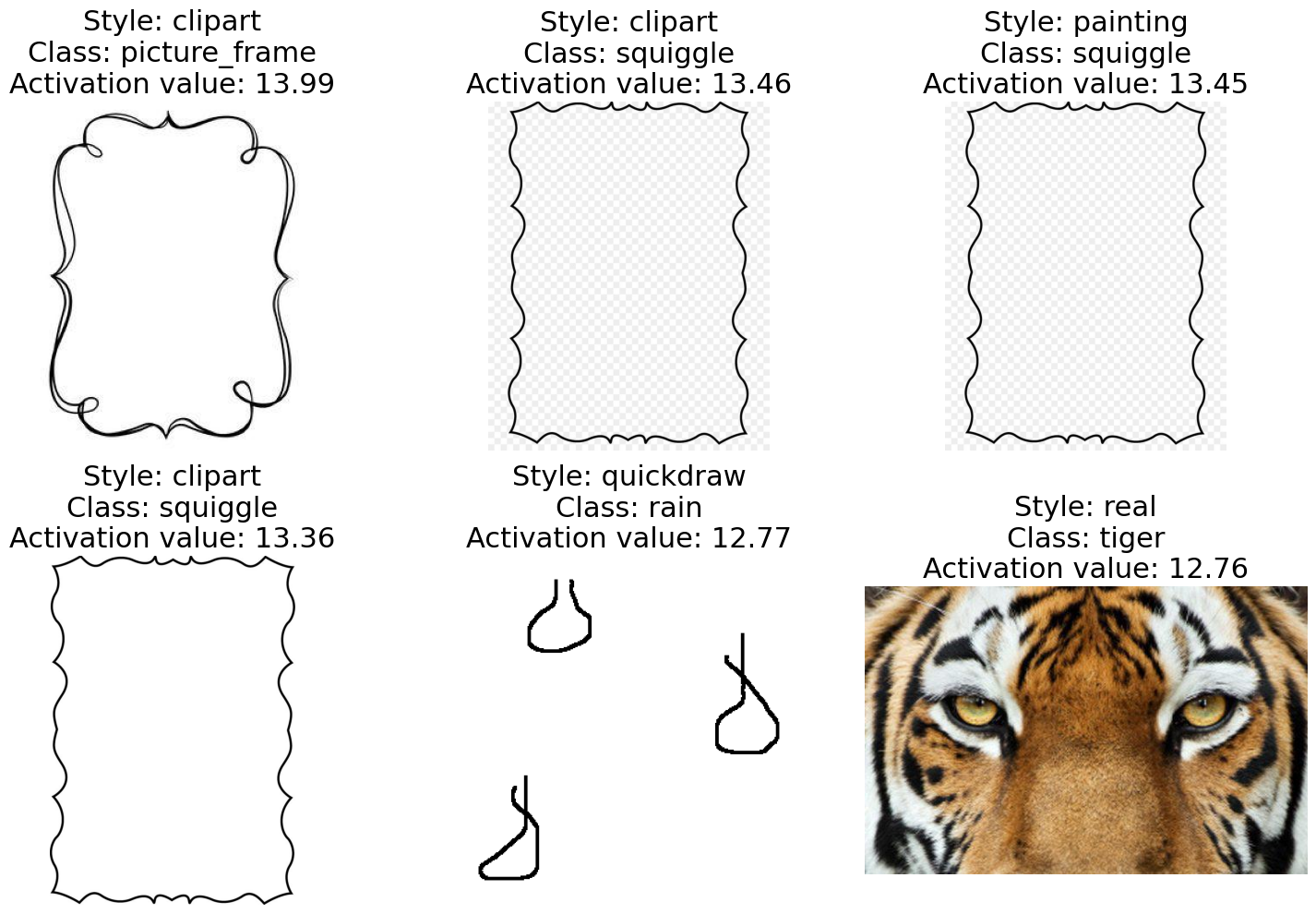} % Replace with your figure file
        \caption{DomainNet, Feature 189}
        \label{fig:sub1}
    \end{subfigure}
    \hfill
    \begin{subfigure}{0.38\textwidth}
        \centering
        \includegraphics[width=\textwidth]{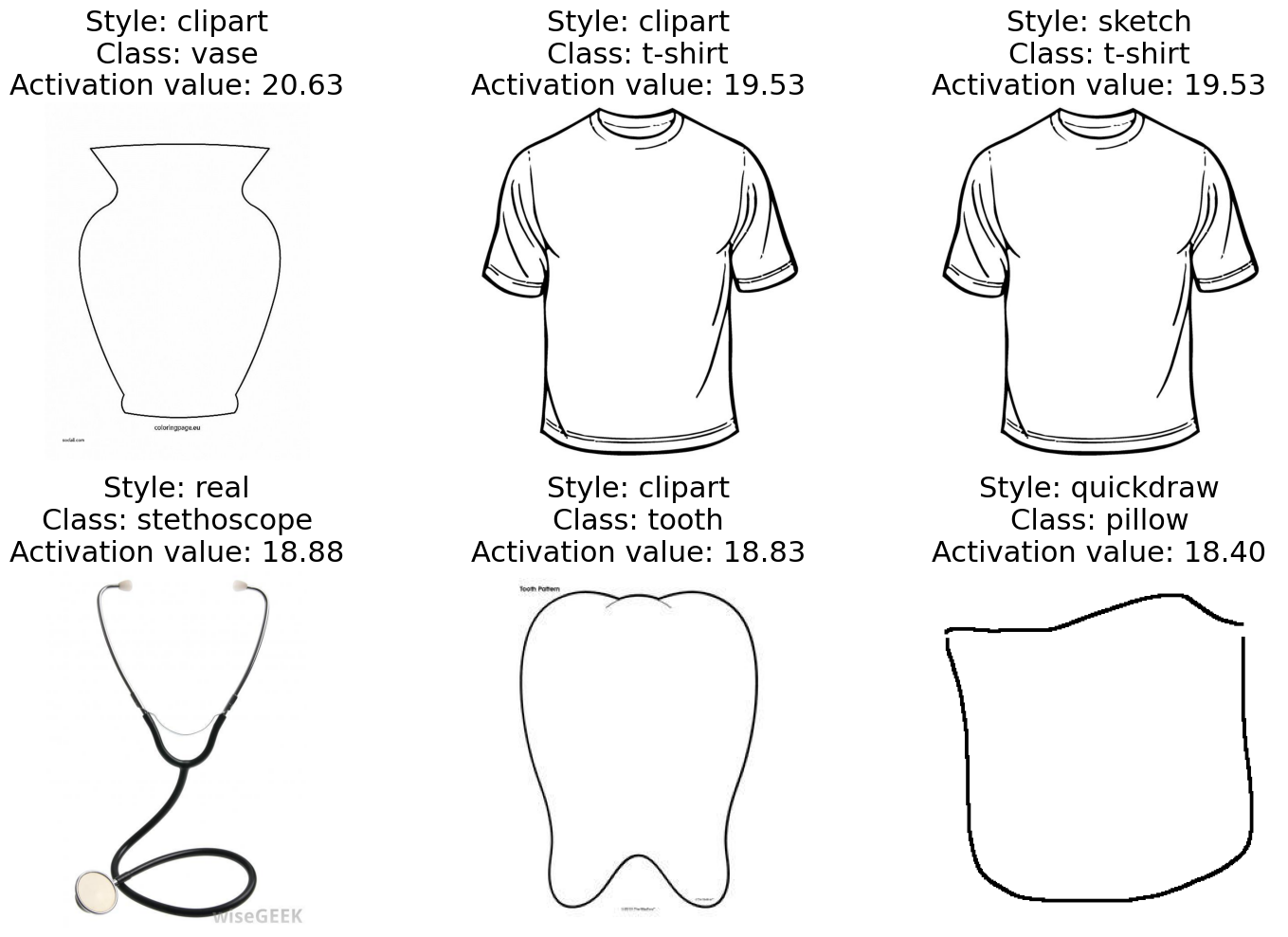} % Replace with your figure file
        \caption{DomainNet, Feature 361}
        \label{fig:sub2}
    \end{subfigure}

    \vskip\baselineskip
    
    \begin{subfigure}{0.38\textwidth}
        \centering
        \includegraphics[width=\textwidth]{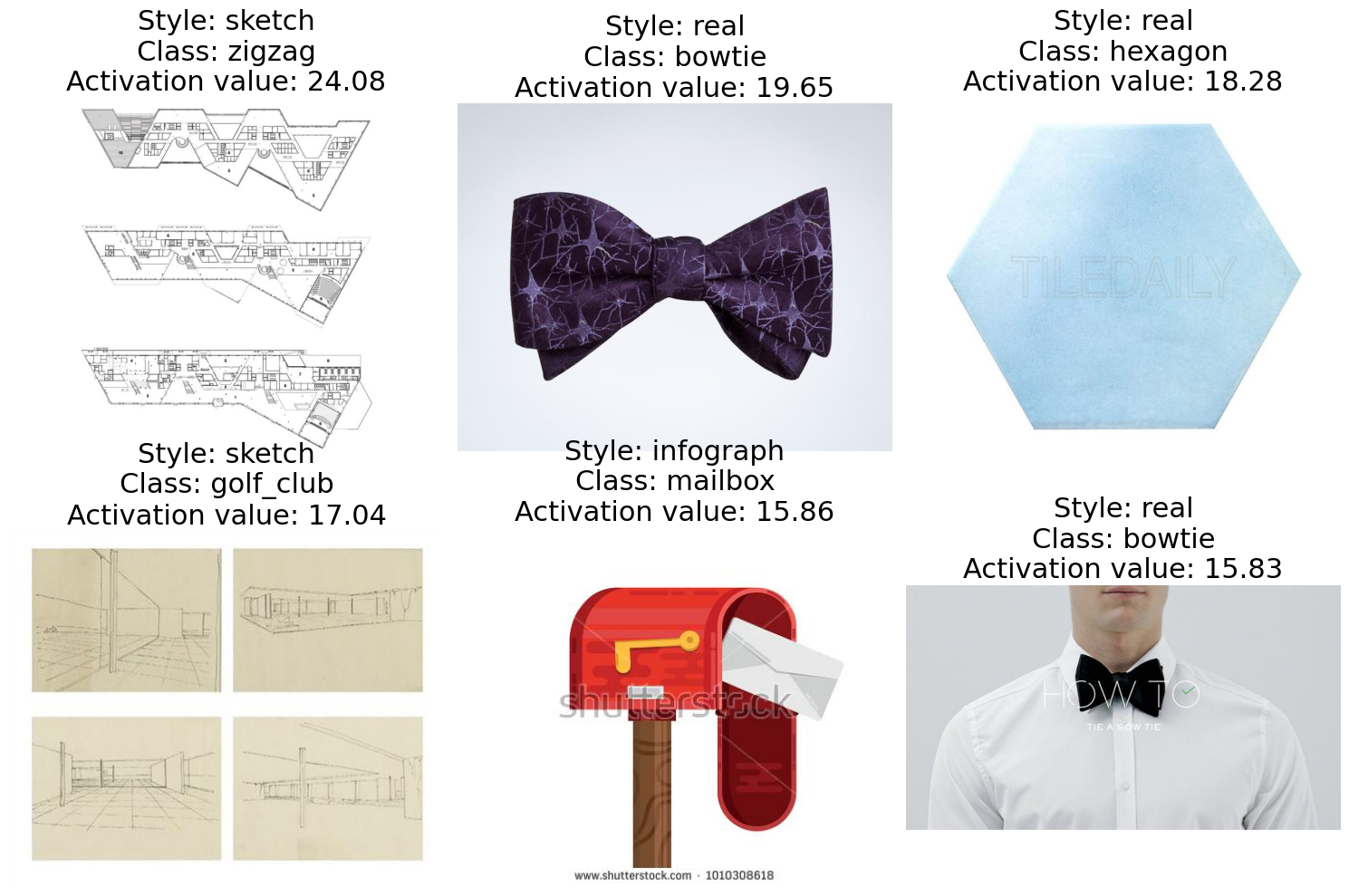} % Replace with your figure file
        \caption{DomainNet, Feature 154}
        \label{fig:sub3}
    \end{subfigure}
    \hfill
    \begin{subfigure}{0.38\textwidth}
        \centering
        \includegraphics[width=\textwidth]{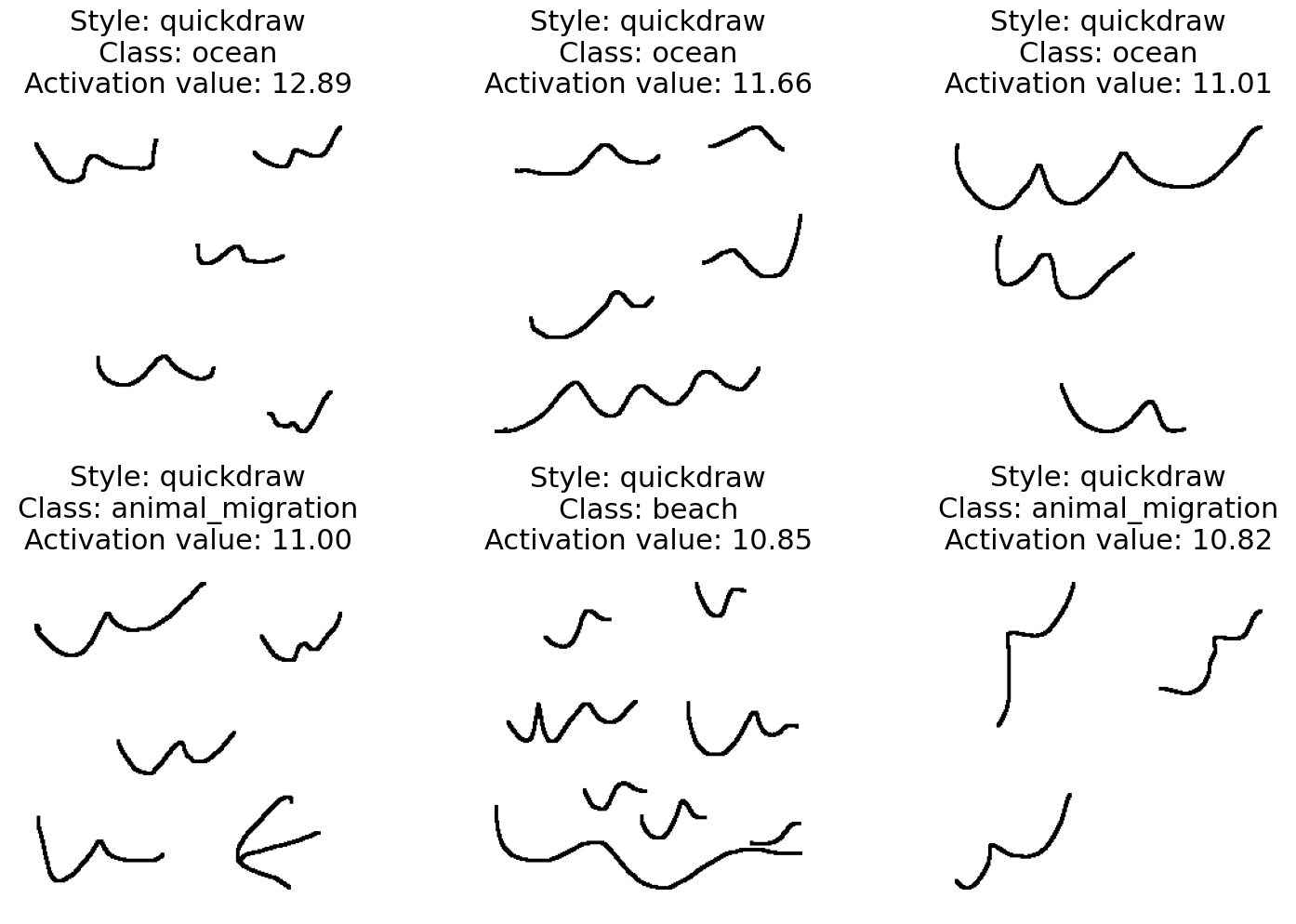} % Replace with your figure file
        \caption{DomainNet, Feature 58}
        \label{fig:sub4}
    \end{subfigure}

    \vskip\baselineskip

    \begin{subfigure}{0.38\textwidth}
        \centering
        \includegraphics[width=\textwidth]{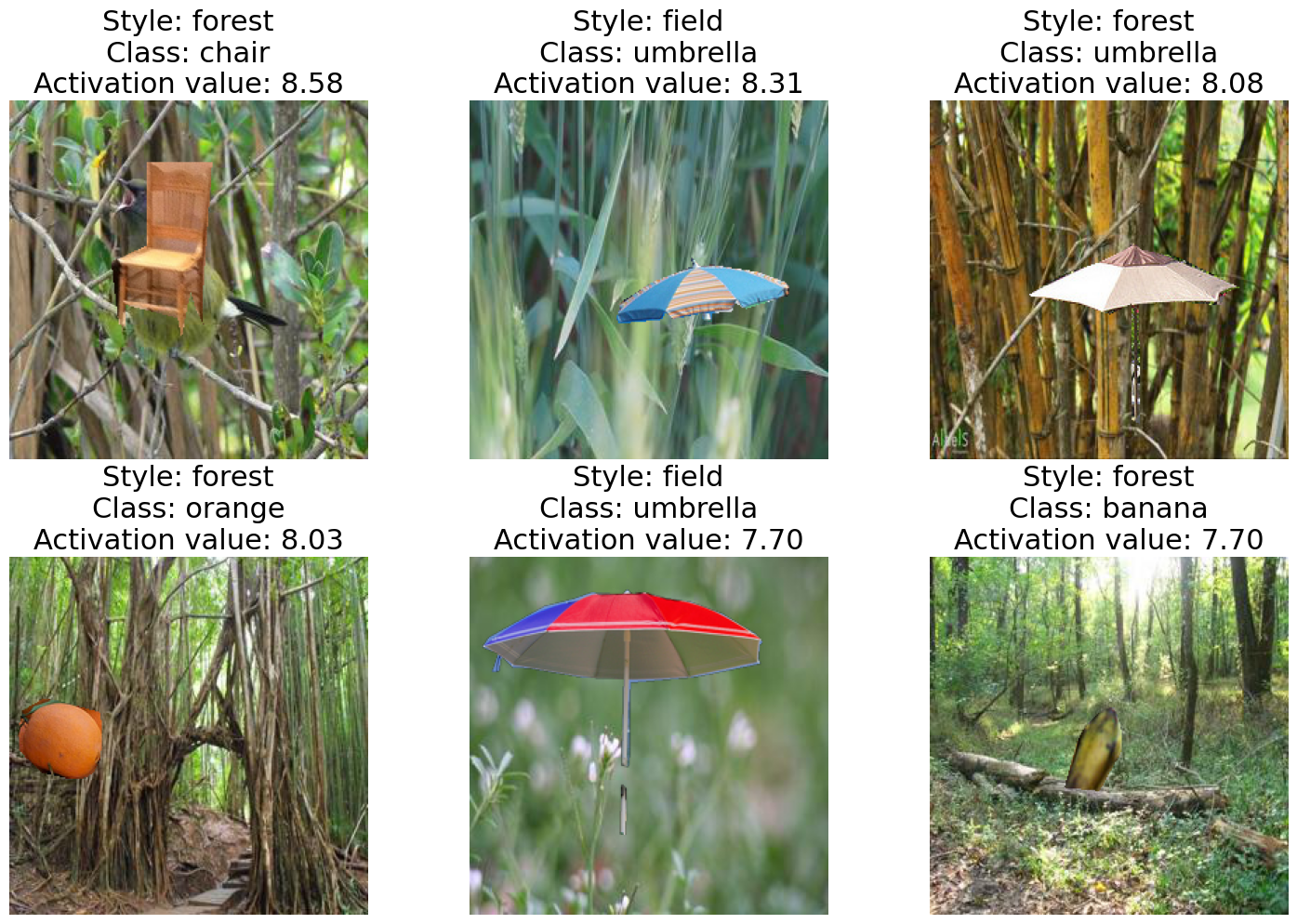} % Replace with your figure file
        \caption{COCOShift95, Feature 297}
        \label{fig:sub5}
    \end{subfigure}
    \hfill
    \begin{subfigure}{0.38\textwidth}
        \centering
        \includegraphics[width=\textwidth]{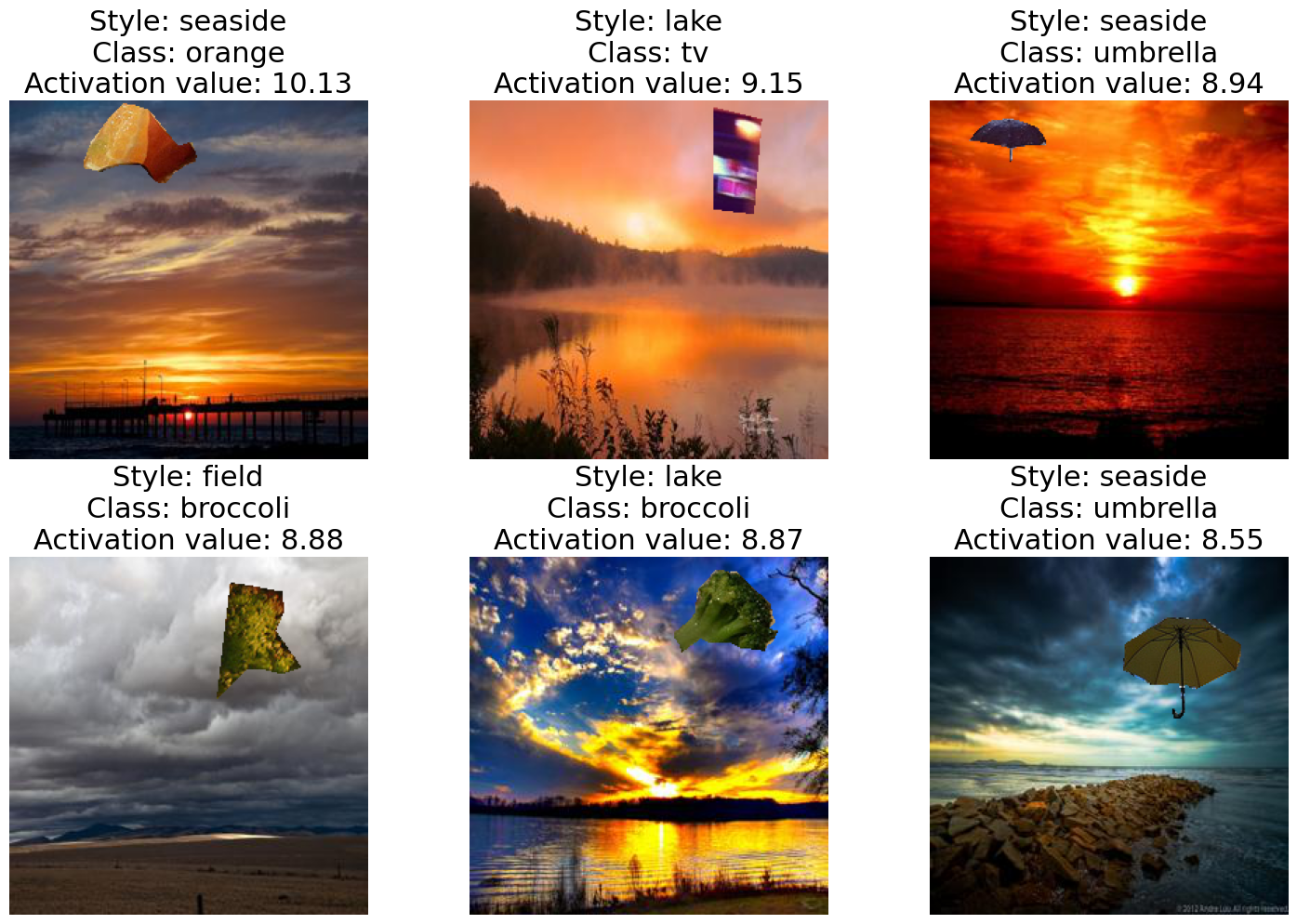} % Replace with your figure file
        \caption{COCOShift95, Feature 435}
        \label{fig:sub6}
    \end{subfigure}

    \vskip\baselineskip
    
    \begin{subfigure}{0.38\textwidth}
        \centering
        \includegraphics[width=\textwidth]{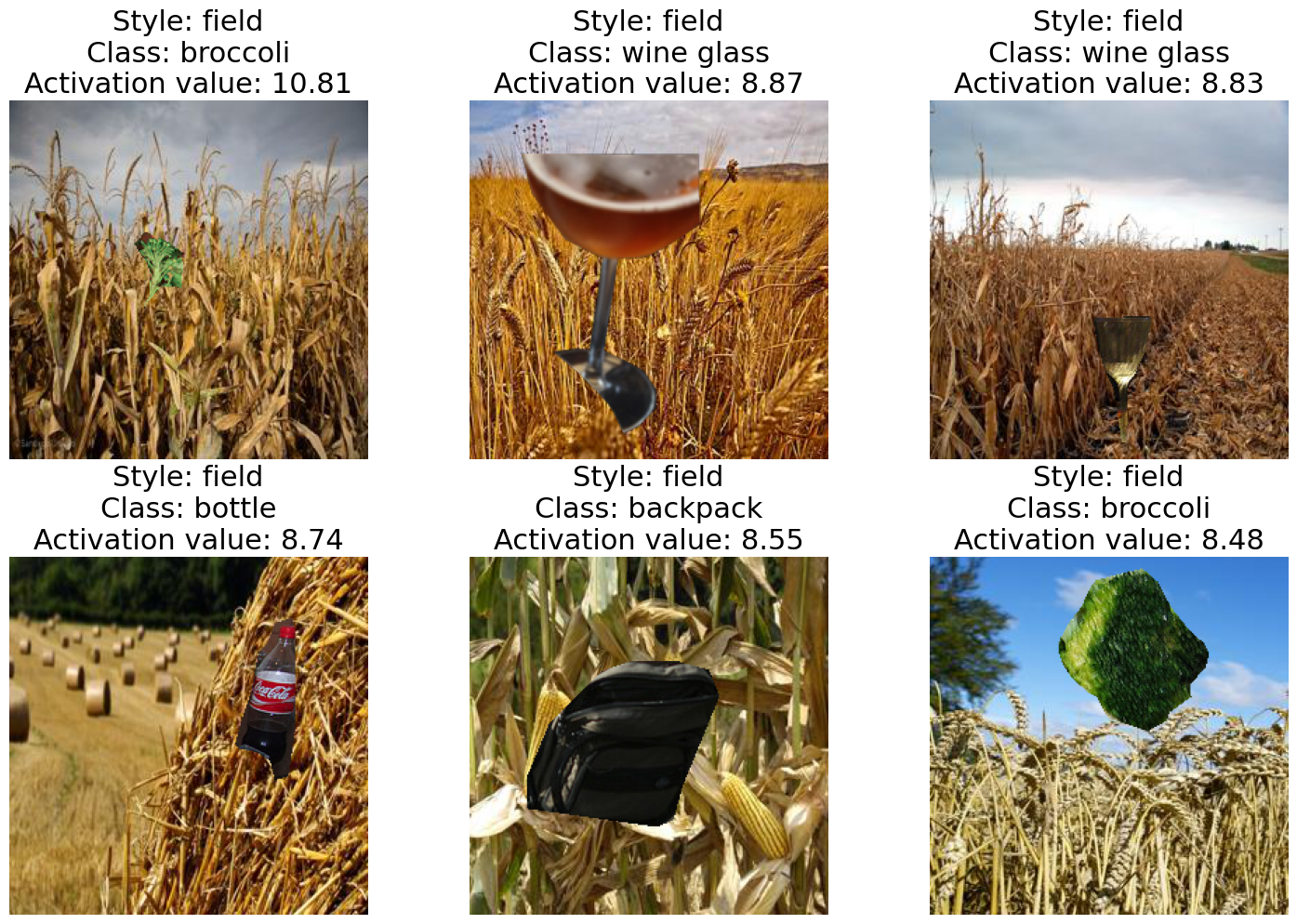} % Replace with your figure file
        \caption{COCOShift95, Feature 252}
        \label{fig:sub7}
    \end{subfigure}

    \caption{Images with the highest activation for a given feature and training dataset (ResNet-18 Embeddings)}
    \label{fig:qualitative_analysis}
\end{figure}

In this section, we analyze the top-ranked features identified by Stylist. Specifically, we extract samples with the highest activation for a given feature and investigate their common characteristics. To this end, Figure~\ref{fig:qualitative_analysis} showcases the top six samples with the highest activations for a selection of features from the top of the Stylist ranking. Each subfigure reveals that the common traits among the images are related to style, particularly centered around environments (sketches/quickdraw for DomainNet and forest/field for COCOShift95). Table~\ref{tab:qualitative_analysis} presents the features alongside their positions in Stylist's ranking, as well as in other ranking algorithms. The table illustrates that some features, highly ranked by Stylist, receive lower ratings from other algorithms. However, the images in Figure~\ref{fig:qualitative_analysis} suggest that these features are indeed related to style.

\begin{table}[ht]
\centering
\caption{The proportion within which features are located for different rankings and training datasets. Features top-ranked by Stylist are ranked lower by InfoGain and PCA Loadings ranking algorithms.}
\begin{tabular}{llccc}
\toprule
\textbf{Training Dataset} & \textbf{Feature Index} & \multicolumn{3}{c}{\textbf{Feature in the Top \% of Rankings}} \\
\cmidrule(lr){3-5}
 &  & \textbf{Stylist} & \textbf{InfoGain} & \textbf{PCA Loadings} \\
\midrule
DomainNet & 189  & 3.12\%   & 17.58\%  & 6.25\%  \\
DomainNet & 361  & 3.71\%   & 29.30\%  & 3.91\%  \\
DomainNet & 154  & 3.91\%   & 8.59\%   & 54.49\% \\
DomainNet & 58   & 4.10\%   & 10.55\%  & 21.09\% \\
\midrule
COCOShift95 & 297 & 0.78\%  & 2.15\%   & 23.83\% \\
COCOShift95 & 435 & 2.34\%  & 31.45\%  & 3.91\%  \\
COCOShift95 & 252 & 2.73\%  & 20.31\%  & 54.69\% \\
\bottomrule
\label{tab:qualitative_analysis}
\end{tabular}
\end{table}

\section{Determine the proportion of selected features}
We can determine the optimal proportion of features following a validation methodology on the ID validation set or by employing an oracle strategy and interrogating the OOD test set. In Tab.~\ref{tab:appendix_cutoff}  we compare these approaches, highlighting that the optimal number of features is stable between the two setups. The variance between the two approaches is less than $0.015$. Also, we emphasize that Stylist always improves over the baseline considering $100\%$ of the features, in all the considered configurations.

\begin{table}[t]
    \centering
    % \captionsetup{\textwidth}
    \caption{Feature selection approach: based on ID validset vs OOD testset. Notice extremely small variances in results, between the two approaches.}
    \setlength{\tabcolsep}{6pt}
    \begin{tabular}{ll ccc ccc ccc}
    \toprule
    &\parbox[t]{0mm}{\multirow{5}{*}{\bf \shortstack{Selection\\Method}}}& \multicolumn{3}{c}{\textbf{COCOShift\_balanced}} & \multicolumn{3}{c}{\textbf{COCOShift75}} & \multicolumn{3}{c}{\textbf{COCOShift95}} \\
    \cmidrule(lr){3-5}
    \cmidrule(lr){6-8}
    \cmidrule(lr){9-11}
    && \multicolumn{2}{c}{ROC-AUC $\uparrow$} & \multirow{2}{*}{\shortstack{\% \\selected\\feat.}} & \multicolumn{2}{c}{ROC-AUC $\uparrow$ } & \multirow{2}{*}{\shortstack{\%\\selected\\feat.}} & \multicolumn{2}{c}{ROC-AUC $\uparrow$ } & \multirow{2}{*}{\shortstack{\%\\selected\\feat.}}\\
    \cmidrule(lr){3-4}
    \cmidrule(lr){6-7}
    \cmidrule(lr){9-10}
    && \shortstack{all\\feat.} & \shortstack{\textbf{kept}\\ feat.} & &  \shortstack{all\\feat.} & \shortstack{\textbf{kept}\\ feat.} & & \shortstack{all\\feat.} & \shortstack{\textbf{kept}\\ feat.}\\ 
    \toprule
    % \parbox[t]{2mm}{\multirow{5}{*}{\rotatebox[origin=c]{90}{\textbf{Env-Aware}}}} 
    \multicolumn{2}{l}{\textbf{ResNet-18}} \\
    \multicolumn{2}{l}{Based on ID val set} & 80.93	& 83.24 ({\footnotesize{\textcolor{ForestGreen}{+2.31}}})& 	10 & 80.40	& 84.71 ({\footnotesize{\textcolor{ForestGreen}{+4.31}}})& 	10 & 79.82	& 85.01 ({\footnotesize{\textcolor{ForestGreen}{+5.19}}})	& 25\\
    \multicolumn{2}{l}{Based on OOD test set} & 80.93 &	83.49 ({\footnotesize{\textcolor{ForestGreen}{+2.56}}}) & 40 & 80.40 &	84.71 ({\footnotesize{\textcolor{ForestGreen}{+4.31}}}) &	10 & 	79.82	& 85.13 ({\footnotesize{\textcolor{ForestGreen}{+5.31}}}) &	30\\
    \multicolumn{2}{c}{\bf Variance} && \bf 0.015 &&& \bf 0.00 &&& \bf 0.003\\
    \midrule
    \multicolumn{2}{l}{\textbf{CLIP}} \\
    \multicolumn{2}{l}{Based on ID val set} &95.06& 	95.40 ({\footnotesize{\textcolor{ForestGreen}{+0.34}}}) & 95 & 94.88 & 95.21 ({\footnotesize{\textcolor{ForestGreen}{+0.33}}}) & 95 & 94.53&94.92 ({\footnotesize{\textcolor{ForestGreen}{+0.39}}}) &95\\
    \multicolumn{2}{l}{Based on OOD test set} &95.06& 	95.40 ({\footnotesize{\textcolor{ForestGreen}{+0.34}}}) & 95 & 94.88 & 95.21 ({\footnotesize{\textcolor{ForestGreen}{+0.33}}}) & 95 & 94.53&94.92 ({\footnotesize{\textcolor{ForestGreen}{+0.39}}}) &95
    \\
    \multicolumn{2}{c}{\bf Variance}&& \bf 0.00 &&& \bf 0.00 &&& \bf 0.00\\
    \bottomrule
    \end{tabular}
    \label{tab:appendix_cutoff}
\end{table}

\section{Future Work}
We leave here several unexplored directions we consider valuable for further investigations:
\begin{enumerate}
    \item Analyze the impact of the pretrained feature extractor, looking after different axes of variation: supervised/unsupervised pretraining, high/low disentanglement. And going even further, find an unsupervised way to choose the best feature extractor, given a dataset. Also, target methods that promise to disentangle the features, like Sparse AE.
    \item Explore more complex algorithms for ranking, based on the same principle of emphasizing the intra and inter environment distances. More related to the algorithm, explore an unsupervised manner to choose the best percent of features to keep. 
    \item Take the approach beyond novelty detection, analyzing the performance improvement of feature ranking and selection w.r.t. other supervised approaches for OOD robustness.
    \item Making the approach more automatic and less environment label dependent, by providing (or couple it with) an environment discovery solution.
    
\end{enumerate}

% \subsection{Qualitative examples}
% Fig.~\ref{fig:qualitative_examples}

% \begin{figure*}[t!]
%     \centering
%     \includegraphics[width=0.7\textwidth]{figures2/qualitative_examples.png}
%     \caption{.}
%     \label{fig:qualitative_examples}
% \end{figure*}

\section{Benchmarks}
\label{appendix:benchmarks}
\subsection{fMoW}\label{appendix:fmow}

\begin{itemize}
    \item Content (functional purpose of buildings):
    \begin{itemize}
        \item {\bf normal}: airport, airport terminal, barn, burial site, car dealership, dam, debris or rubble, educational institution, electric substation, fountain, gas station, golf course, hospital, interchange, multi-unit residential, parking lot or garage, police station, port, railway bridge, recreational facility, road bridge, runway, shipyard, shopping mall, solar farm, space facility, surface mine, swimming pool, waste disposal, water treatment facility, zoo
        \item {\bf novel}: airport hangar, amusement park, aquaculture, archaeological site, border checkpoint, construction site, crop field, factory or powerplant, fire station, flooded road, ground transportation station, helipad, impoverished settlement, lake or pond, lighthouse, military facility, nuclear powerplant, office building, oil or gas facility, park, place of worship, prison, race track, single-unit residential, smokestack, stadium, storage tank, toll booth, tower, tunnel opening, wind farm
    \end{itemize}
    \item Style (geographical area):
    \begin{itemize}
        \item {\bf ID}: Europe, America, Asia, Africa
        \item {\bf OOD}: Australia
    \end{itemize}
\end{itemize}

\begin{itemize}
    \item Number of samples:
    \begin{itemize}
        \item OOD test set: 3469
        % train + valid
        \item ID train set: 55288
        % test
        \item ID test set: 13817
        % valid
        \item ID val set: 6911
    \end{itemize}
\end{itemize}

 % \stefan{esti sigur? aici aveam 4 as OOD, care cred ca era oceania?!}
 % Normals: ['airport', 'airport_terminal', 'barn', 'burial_site', 'car_dealership', 'dam' ,'debris_or_rubble', 'educational_institution', 'electric_substation', 'fountain', 'gas_station', 'golf_course', 'hospital', 'interchange', 'multi-unit_residential', 'parking_lot_or_garage', 'police_station', 'port', 'railway_bridge', 'recreational_facility', 'road_bridge', 'runway' ,'shipyard' ,'shopping_mall', 'solar_farm', 'space_facility', 'surface_mine', 'swimming_pool', 'waste_disposal', 'water_treatment_facility', 'zoo']
 % Novels:['airport_hangar', 'amusement_park', 'aquaculture', 'archaeological_site', 'border_checkpoint', 'construction_site', 'crop_field', 'factory_or_powerplant', 'fire_station', 'flooded_road', 'ground_transportation_station', 'helipad' ,'impoverished_settlement', 'lake_or_pond', 'lighthouse', 'military_facility', 'nuclear_powerplant', 'office_building', 'oil_or_gas_facility', 'park', 'place_of_worship' ,'prison', 'race_track', 'single-unit_residential', 'smokestack', 'stadium' ,'storage_tank' ,'toll_booth', 'tower', 'tunnel_opening' ,'wind_farm']
 
\subsection{DomainNet}\label{appendix:domainnet}

    \begin{itemize}
        \item Content (object classes):
        \begin{itemize}
            \item {\bf normal}: aircraft carrier, angel, animal migration, apple, arm, backpack, barn, basketball, bed, belt, birthday cake, blackberry, blueberry, book, boomerang, bowtie, brain, bread, bucket, butterfly, cactus, cake, camouflage, cannon, carrot, cat, cello, chandelier, circle, cloud, coffee cup, computer, cookie, couch, crab, crayon, crocodile, cruise ship, diamond, diving board, dog, donut, door, dresser, drill, drums, duck, ear, elbow, envelope, eraser, fan, fence, flower, flying saucer, fork, frog, garden, guitar, hand, headphones, helicopter, helmet, hexagon, hockey stick, horse, hospital, hot air balloon, hot dog, hourglass, house, hurricane, jacket, jail, kangaroo, knife, laptop, leg, light bulb, lighter, lightning, lipstick, lobster, map, microphone, microwave, mountain, moustache, mug, mushroom, necklace, nose, owl, paint can, paintbrush, palm tree, parachute, parrot, peanut, pear, peas, piano, pig, pillow, pineapple, pizza, police car, pond, postcard, power outlet, radio, rain, rake, remote control, roller coaster, sailboat, saw, saxophone, screwdriver, sea turtle, see saw, shark, shorts, skull, sleeping bag, snail, snowman, soccer ball, spider, spoon, square, stairs, star, stethoscope, stitches, stop sign, strawberry, streetlight, string bean, submarine, suitcase, sun, swan, sweater, swing set, syringe, table, teapot, teddy-bear, telephone, television, tennis racquet, tent, toaster, toe, tooth, traffic light, train, tree, trombone, truck, trumpet, umbrella, underwear, van, vase, violin, whale, wheel, wine bottle, wristwatch, zebra
        
            \item {\bf novel}: airplane, alarm clock, ambulance, ant, anvil, asparagus, axe, banana, bandage, baseball, baseball bat, basket, bat, bathtub, beach, bear, beard, bee, bench, bicycle, binoculars, bird, bottlecap, bracelet, bridge, broccoli, broom, bulldozer, bus, bush, calculator, calendar, camel, camera, campfire, candle, canoe, car, castle, ceiling fan, cell phone, chair, church, clarinet, clock, compass, cooler, cow, crown, cup, dishwasher, dolphin, dragon, dumbbell, elephant, eye, eyeglasses, face, feather, finger, fire hydrant, fireplace, firetruck, fish, flamingo, flashlight, flip flops, floor lamp, foot, frying pan, garden hose, giraffe, goatee, golf club, grapes, grass, hamburger, hammer, harp, hat, hedgehog, hockey puck, hot tub, house plant, ice cream, key, keyboard, knee, ladder, lantern, leaf, lighthouse, line, lion, lollipop, mailbox, marker, matches, megaphone, mermaid, monkey, moon, mosquito, motorbike, mouse, mouth, nail, ocean, octagon, octopus, onion, oven, panda, pants, paper clip, passport, pencil, penguin, pickup truck, picture frame, pliers, pool, popsicle, potato, purse, rabbit, raccoon, rainbow, rhinoceros, rifle, river, rollerskates, sandwich, school bus, scissors, scorpion, sheep, shoe, shovel, sink, skateboard, skyscraper, smiley face, snake, snorkel, snowflake, sock, speedboat, spreadsheet, squiggle, squirrel, steak, stereo, stove, sword, t-shirt, The Eiffel Tower, The Great Wall of China, The Mona Lisa, tiger, toilet, toothbrush, toothpaste, tornado, tractor, triangle, washing machine, watermelon, waterslide, windmill, wine glass, yoga, zigzag
        \end{itemize}
        \item Style (manner of depiction):
        \begin{itemize}
            \item {\bf ID}: real, painting, clipart, infograph
            \item {\bf OOD}: sketch, quickdraw
        \end{itemize}
    \end{itemize}

    \begin{itemize}
        \item Number of samples:
        \begin{itemize}
            \item OOD test set: 242886
            % train + valid
            \item ID train set: 142026
            % test 
            \item ID test set: 35313
            % valid 
            \item ID val set: 17753
        \end{itemize}
    \end{itemize}

\subsection{COCOShift}\label{appendix:cocoshift}
Each COCOShift environment is composed of 5 closely related categories of Places365 as follows:

\begin{itemize}
    \item {\bf forest}: forest, rainforest, bamboo\_forest, forest\_path, forest\_road,
    \item {\bf mountain}: mountain, mountain\_snowy, glacier, mountain\_path, crevasse
    \item {\bf seaside}: beach, coast, ocean, boathouse, beach\_house"
    \item {\bf garden}: botanical\_garden, formal\_garden, japanese\_garden, vegetable\_garden, greenhouse
    \item {\bf field}: field\_cultivated, field\_wild, wheat\_field, corn\_field, field\_road
    \item {\bf rock}: badlands, butte, canyon, cliff, grotto
    \item {\bf lake}: lake, lagoon, swamp, marsh, hot\_spring
    \item {\bf farm}: orchard, vineyard, farm, rice\_paddy, pasture
    \item {\bf{sport\_field}}: soccer\_field, football\_field, golf\_course, baseball\_field, athletic\_field
\end{itemize}

On the other hand, we worked with superclasses from COCO~\cite{COCO} such that there would be a significant shift between classes. 

To make sure that the content is identifiable from each image (as many COCO segmentations provide little content without it's context), we tested each generated image against CLIP~\cite{clip}. Specifically, we took a given merged picture into COCOShift dataset if CLIP could correctly identify the content between the selected COCO classes (not superclasses) listed below. The same test is effectuated on images of segmentations over white backgrounds.
    % Places envs are aggregation of more: ...
    % COCO segmentations supercategories:
    
    \begin{itemize}
        \item Content (object category):
        \begin{itemize}
            \item {\bf normal}: food (composed of classes: hot dog, cake, donut, carrot, sandwich, broccoli, banana, apple, pizza, orange)
            \item {\bf novel}: electronic (composed of classes: remote, laptop, tv, cell phone, keyboard), kitchen (composed of classes: bottle, cup, wine glass, knife, fork, bowl, spoon)
        \end{itemize}
        \item Style (background area surrounding the object):
        \begin{itemize}
            \item {\bf ID}: forest, mountain, field, rock, farm
            \item {\bf OOD}: lake, seaside, garden, sport field
        \end{itemize}
       
    \end{itemize}

    \begin{itemize}
        \item Number of samples:
        \begin{itemize}
            \item OOD test set: 13013
            \item ID test set: 1623
            % train + valid
            \item COCOShift\_balanced 
                \begin{itemize}
                    \item ID train set: 7033
                    \item ID val set: 880
                \end{itemize}
            \item COCOShift75 
                \begin{itemize}
                    \item ID train set: 4221
                    \item ID val set: 529
                \end{itemize}
            \item COCOShift90 
                \begin{itemize}
                    \item ID train set: 3284
                    \item ID val set: 412
                \end{itemize}
            \item COCOShift95
                \begin{itemize}
                    \item ID train set: 3037
                    \item ID val set: 379
                \end{itemize}
        \end{itemize}
    \end{itemize}

    \begin{itemize}
        \item {\bf Spurious correlation} sets are variants of train sets used for the synthetic COCOShift dataset, where we eliminate samples to create spuriousity.
        \begin{itemize}
            \item {\bf COCOShift\_balanced}: normal samples are uniformly distributed among the ID environments ($\approx1.4k$ samples per environment)
            \item {\bf COCOShift75}: environments [farm, mountain] have $\approx1.4k$ normal samples, while environments [rock, forest, field] have $\approx400$ normal samples
            \item {\bf COCOShift90}: environments [farm, mountain] have $\approx1.4k$ normal samples, while environments [rock, forest, field] have $\approx150$ normal samples
            \item {\bf COCOShift95}: environments [farm, mountain] have $\approx1.4k$ normal samples, while environments [rock, forest, field] have $\approx70$ normal samples
        \end{itemize}
    \end{itemize}
\end{document}